\documentclass{article}

\usepackage{microtype}
\usepackage{graphicx}
\usepackage{subfigure}
\usepackage{booktabs} % for professional tables

\usepackage{hyperref}

\usepackage{url}
\usepackage{amsfonts}
\usepackage{nicefrac}
\usepackage{multicol}
\usepackage{tikz}
\usepackage{diagbox}
\usepackage{multirow}
\usepackage{caption}
\usepackage{subcaption}
\usepackage{bbm}
\usepackage{bm}
\usepackage{amsmath}
\usepackage{amsthm}
\usepackage{amssymb}
\usepackage{bm}
\usepackage{textcomp}
\usepackage{footnote}
\usepackage{wrapfig}
\usepackage[export]{adjustbox}
\usepackage{notation}
\usepackage{researchpack}
\usepackage{algorithm}
\usepackage[noend]{algorithmic}
\usepackage{tikz}
\usetikzlibrary{tikzmark}

\usepackage[capitalize,noabbrev]{cleveref}

\crefname{defn}{Definition}{Definition}
\crefname{section}{Section}{Section}
\crefname{algorithm}{Algorithm}{Algorithm} 
\crefname{thm}{Thm.}{Thm.}
\crefname{lem}{Lemma}{Lemma}
\crefname{prop}{Prop.}{Prop.}
\crefname{asm}{Asm.}{Asm.}
\crefname{appendix}{Appendix}{Appx.}
\crefname{equation}{Equation}{Equations}
\crefname{figure}{Figure}{Figure}
\crefname{table}{Table}{Table}
\crefname{cor}{Corollary}{Corollary}

\usepackage[accepted]{icml2024}

\usepackage{amsmath}
\usepackage{amssymb}
\usepackage{mathtools}
\usepackage{amsthm}

\usepackage[capitalize,noabbrev]{cleveref}

\theoremstyle{plain}

\theoremstyle{definition}

\theoremstyle{remark}

\newcolumntype{R}{>{}r<{}}
\newcolumntype{L}{>{}l<{}}
\newcolumntype{M}{R@{}L}

\usepackage[textsize=tiny]{todonotes}

\icmltitlerunning{Scaling Tractable Probabilistic Circuits: A Systems Perspective}

\begin{document}

\twocolumn[
\icmltitle{Scaling Tractable Probabilistic Circuits: A Systems Perspective}



\icmlsetsymbol{equal}{*}

\begin{icmlauthorlist}
\icmlauthor{Anji Liu}{ucla}
\icmlauthor{Kareem Ahmed}{ucla}
\icmlauthor{Guy Van den Broeck}{ucla}
\end{icmlauthorlist}

\icmlaffiliation{ucla}{Department of Computer Science, University of California, Los Angeles, USA}

\icmlcorrespondingauthor{Anji Liu}{liuanji@cs.ucla.edu}

\icmlkeywords{Machine Learning, ICML}

\vskip 0.3in
]



\printAffiliationsAndNotice{}  % leave blank if no need to mention equal contribution

\begin{abstract}
Probabilistic Circuits (PCs) are a general framework for tractable deep generative models, which support exact and efficient probabilistic inference on their learned distributions. Recent modeling and training advancements have enabled their application to complex real-world tasks. However, the time and memory inefficiency of existing PC implementations hinders further scaling up. This paper proposes PyJuice, a general GPU implementation design for PCs that improves prior art in several regards. Specifically, PyJuice is 1-2 orders of magnitude faster than existing systems (including very recent ones) at training large-scale PCs. Moreover, PyJuice consumes 2-5x less GPU memory, which enables us to train larger models. At the core of our system is a compilation process that converts a PC into a compact representation amenable to efficient block-based parallelization, which significantly reduces IO and makes it possible to leverage Tensor Cores available in modern GPUs. Empirically, PyJuice can be used to improve state-of-the-art PCs trained on image (\eg ImageNet32) and language (\eg WikiText, CommonGen) datasets. We further establish a new set of baselines on natural image and language datasets by benchmarking existing PC structures but with much larger sizes and more training epochs, with the hope of incentivizing future research. Code is available at \url{https://github.com/Tractables/pyjuice}.
\end{abstract}

% \vspace{-0.5em}
\section{Introduction}
% \vspace{-0.2em}

Many tasks require not only precise modeling of intricate, high-dimensional data distributions but also the efficient execution of probabilistic inference on the learned model. To satisfy inference-side demands, tractable deep generative models are designed to support efficient computation of various probabilistic queries. Probabilistic Circuits (PCs) \citep{choi20probabilistic,vergari2020probabilistic} are a unified framework that abstracts a myriad of tractable model families. PCs have been applied to many domains such as explainability and causality \citep{correia2020joints,wang2023compositional}, graph link prediction \citep{loconte2023turn}, and neuro-symbolic AI \citep{xu2018sl, manhaeve18deepproblog,ahmed2022SPLs}. In particular, there is a trend of using PCs' tractability to control expressive deep generative models, including (large) language models \citep{zhang2023tractable}, image diffusion models \citep{liu2023image}, and reinforcement learning models \citep{liu2023expressive}.

The backbone of the application-side advancements is the recent breakthroughs on the modeling and learning side of PCs, which include designing better PC structures \citep{peharz2020random,correia2023continuous,mathur2023knowledge,loconte2023subtractive,gala2024probabilistic}, effective structure learning algorithms \citep{gens2013learning,dang2020strudel,dang2022sparse,yang2023bayesian}, and distilling from expressive deep generative models \citep{liu2022scaling}. Despite such algorithmic innovations, a fundamental obstacle to further scaling up PC learning and inference is the time and memory inefficiency of existing implementations, hindering the training of large PC models and their application to large-scale datasets.

In this work, we develop an efficient and flexible system called PyJuice that addresses various training and inference tasks for PCs. As shown in \cref{tab:speed-results}, PyJuice is orders of magnitude faster than previous implementations for PCs (\eg SPFlow \citep{molina2019spflow}, EiNet \citep{peharz2020einsum}, and Juice.jl \citep{dang2021juice}) as well as Hidden Markov Models\footnote{Every HMM has an equivalent PC representation.} (\eg Dynamax \citep{murphy2023dynamax}). Additionally, as we shall demonstrate in the experiments, PyJuice is more memory efficient than the baselines, enabling us to train larger PCs with a fixed memory quota.

Unlike other deep generative models based on neural network layers that are readily amenable to efficient systems (\eg a fully connected layer can be emulated by a single matrix multiplication and addition kernel plus an element-wise activation kernel), PCs cannot be \emph{efficiently} computed using well-established operands due to (i) the unique connection patterns of their computation graph,\footnote{Commonly used neural network layers mainly employ ``regular'' tensor operations such as matrix multiplications and tensor inner-/outer-products. In contrast, PC layers can contain nodes that are sparsely connected.} and (ii) the existence of log probabilities at drastically different scales in the models, which requires to properly handle numerical underflow problems. To parallelize PCs at scale, we propose a compilation phase that converts a PC into a compact data structure amenable to block-based parallelization on modern GPUs. Further, we improve the backpropagation process by indirectly computing the parameter updates by backpropagating a quantity called PC flow \citep{choi21group} that is more numerically convenient yet mathematically equivalent.

In the following, we first formally define PCs and discuss common ways to parallelize their computation in \cref{sec:background}. \cref{sec:problems} examines the key bottlenecks in PC parallelization. \cref{sec:blk-sp-parallel,sec:eff-backward} explains our design in details.

\begin{table}[t]
    \centering
    \caption{\textbf{Average ($\pm$ standard deviation of $5$ runs) runtime (in seconds) per training epoch} of 60K samples for PyJuice and the baselines SPFlow \citep{molina2019spflow}, EiNet \citep{peharz2020einsum}, Juice.jl \citep{dang2021juice}, and Dynamax \citep{murphy2023dynamax}. We adopted four PC structures: PD, RAT-SPN, HCLT, and HMM. All experiments were carried out on an RTX 4090 GPU with 24GB memory. To maximize parallelism, we always use the maximum possible batch size. ``OOM'' denotes out-of-memory with batch size $2$. The best numbers are in boldface.}
    \vspace{0.2em}
    \label{tab:speed-results}
    \renewcommand{\arraystretch}{0.88}
    \centering
    \scalebox{0.85}{
    \begin{tabular}{l@{\hspace{0.2em}}M@{\hspace{0.2em}}M@{\hspace{0.2em}}M@{\hspace{0.2em}}M@{\hspace{0.2em}}M}
        \toprule
         & \multicolumn{10}{c}{PD \citep{poon2011sum}} \\
         \cmidrule(lr){2-11}
        \# nodes & 172&K & 344&K & 688&K & 1.38&M & 2.06&M \\
        \# edges & 15.6&M & 56.3&M & 213&M & 829&M & 2.03&B \\
        \midrule
        SPFlow & \multicolumn{2}{c}{$\;\;\,$\footnotesize{$>\!\!25000$}} & \multicolumn{2}{c}{$\!$\footnotesize{$>\!\!25000$}} & \multicolumn{2}{c}{$\!$\footnotesize{$>\!\!25000$}} & \multicolumn{2}{c}{$\!\!\!$\footnotesize{$>\!\!25000$}} & \multicolumn{2}{c}{$\!\!\!$\footnotesize{$>\!\!25000$}} \\
        EiNet & 34.2&{\tiny$\pm$0.0} & 88.7&{\tiny$\pm$0.2} & 456.1&{\tiny$\pm$2.3} & 1534.7&{\tiny$\pm$0.5} & \multicolumn{2}{c}{$\!\!\!$OOM} \\
        Juice.jl & 12.6&{\tiny$\pm$0.5} & 37.0&{\tiny$\pm$1.7} & 141.7&{\tiny$\pm$6.9} & \multicolumn{2}{c}{$\!\!\!$OOM} & \multicolumn{2}{c}{$\!\!\!$OOM} \\
        PyJuice & \textbf{2.0}&{\tiny$\pm$0.0} & \textbf{5.3}&{\tiny$\pm$0.0} & \textbf{15.4}&{\tiny$\pm$0.0} & \textbf{57.1}&{\tiny$\pm$0.2} & \textbf{203.7}&{\tiny$\pm$0.1} \\
        \midrule
         & \multicolumn{10}{c}{RAT-SPN \citep{peharz2020random}} \\
        \cmidrule(lr){2-11}
        \# nodes & 58&K & 116&K & 232&K & 465&K & 930&K \\
        \# edges & 616&K & 2.2&M & 8.6&M & 33.4&M & 132&M \\
        \midrule
        SPFlow & 6372.1&{\tiny$\pm$4.2} & \multicolumn{2}{c}{$\!$\footnotesize{$>\!\!25000$}} & \multicolumn{2}{c}{$\!$\footnotesize{$>\!\!25000$}} & \multicolumn{2}{c}{$\!$\footnotesize{$>\!\!25000$}} & \multicolumn{2}{c}{$\!\!\!$\footnotesize{$>\!\!25000$}} \\
        EiNets & 38.5&{\tiny$\pm$0.0} & 83.5&{\tiny$\pm$0.0} & 193.5&{\tiny$\pm$0.1} & 500.6&{\tiny$\pm$0.2} & 2445.1&{\tiny$\pm$2.6} \\
        Juice.jl & 6.0&{\tiny$\pm$0.3} & 9.4&{\tiny$\pm$0.3} & 25.5&{\tiny$\pm$2.4} & 84.0&{\tiny$\pm$4.0} & 375.1&{\tiny$\pm$3.4} \\
        PyJuice & \textbf{0.6}&{\tiny$\pm$0.0} & \textbf{0.9}&{\tiny$\pm$0.1} & \textbf{1.6}&{\tiny$\pm$0.0} & \textbf{5.8}&{\tiny$\pm$0.1} & \textbf{13.8}&{\tiny$\pm$0.0} \\
        \midrule
         & \multicolumn{10}{c}{HCLT \citep{liu2021tractable}} \\
        \cmidrule(lr){2-11}
        \# nodes & 89&K & 178&K & 355&K & 710&K & 1.42&M \\
        \# edges & 2.56&M & 10.1&M & 39.9&M & 159&M & 633&M \\
        \midrule
        SPFlow & 22955.6&{\tiny$\pm$18.4} & \multicolumn{2}{c}{$\!$\footnotesize{$>\!\!25000$}} & \multicolumn{2}{c}{$\!$\footnotesize{$>\!\!25000$}} & \multicolumn{2}{c}{$\!$\footnotesize{$>\!\!25000$}} & \multicolumn{2}{c}{$\!\!\!$\footnotesize{$>\!\!25000$}} \\
        EiNet & 52.5&{\tiny$\pm$0.3} & 77.4&{\tiny$\pm$0.4} & 233.5&{\tiny$\pm$2.8} & 1170.7&{\tiny$\pm$8.9} & 5654.3&{\tiny$\pm$17.4} \\
        Juice.jl & 4.7&{\tiny$\pm$0.2} & 6.4&{\tiny$\pm$0.5} & 12.4&{\tiny$\pm$1.3} & 41.1&{\tiny$\pm$0.1} & 143.2&{\tiny$\pm$5.1} \\
        PyJuice & \textbf{0.8}&{\tiny$\pm$0.0} & \textbf{1.3}&{\tiny$\pm$0.0} & \textbf{2.6}&{\tiny$\pm$0.0} & \textbf{8.8}&{\tiny$\pm$0.0} & \textbf{24.9}&{\tiny$\pm$0.1} \\
        \midrule
         & \multicolumn{10}{c}{HMM \citep{rabiner1986introduction}} \\
        \cmidrule(lr){2-11}
        \# nodes & 33&K & 66&K & 130&K & 259&K & 388&K \\
        \# edges & 8.16&M & 32.6&M & 130&M & 520&M & 1.17&B \\
        \midrule
        Dynamax & 111.3&{\tiny$\pm$0.4} & 441.2&{\tiny$\pm$3.9} & 934.7&{\tiny$\pm$6.3} & 2130.5&{\tiny$\pm$19.5} & 4039.8&{\tiny$\pm$38.3} \\
        Juice.jl & 4.6&{\tiny$\pm$0.1} & 18.8&{\tiny$\pm$0.1} & 91.6&{\tiny$\pm$0.1} & \multicolumn{2}{c}{$\!\!\!$OOM} & \multicolumn{2}{c}{$\!\!\!$OOM} \\
        PyJuice & \textbf{0.6}&{\tiny$\pm$0.0} & \textbf{1.0}&{\tiny$\pm$0.0} & \textbf{2.9}&{\tiny$\pm$0.1} & \textbf{10.1}&{\tiny$\pm$0.2} & \textbf{39.9}&{\tiny$\pm$0.1} \\
        \bottomrule
    \end{tabular}}
    % \vspace{-0.6em}
\end{table}

% \vspace{-0.4em}
\section{Preliminaries and Related Work}
\label{sec:background}
% \vspace{-0.2em}

% Through the identification of shared structural features,\guy{meh, this is vague. rather say something like this: probabilistic inference is often done by computing sums of products, and PCs just represent the model directly in this way, as a computation graph. There are many bespoke representations of tractable probability distributions, for example SPN, AC, HMM, etc., but they can all be thought of as representing such a computation graph.} 

Many probabilistic inference tasks can be cast into computing sums of products. By viewing them from a computation graph standpoint, PCs provide a unified perspective on many bespoke representations of tractable probability distributions, including Arithmetic Circuits \citep{darwiche2002logical,darwiche2003differential}, Sum-Product Networks \citep{poon2011sum}, Cutset Networks \citep{rahman2014cutset}, and Hidden Markov Models \citep{rabiner1986introduction}. Specifically, PCs define distributions with computation graphs consisting of sum and product operations, as elaborated below.

\begin{defn}[Probabilistic Circuit]
\label{def:pc}
A PC defined over variables $\X$ is represented by a parameterized Directed Acyclic Graph (DAG) with a single root node $n_{\mathrm{r}}$. Every leaf node in the DAG represents an input node that defines a primitive distribution over some variable $X \!\in\! \X$. Every inner node $n$ is either a sum node or a product node, which merges the distributions encoded by its children, denoted $\ch(n)$, to construct more complex distributions. The distribution represented by every node is defined recursively as:
    {\setlength{\abovedisplayskip}{0.4em}
    \setlength{\belowdisplayskip}{-0.2em}
    \begin{align}
        \p_{n} (\x) \!:=\! \begin{cases}
            f_{n} (\x) & n \text{~is~an~input~node}, \\
            \prod_{c \in \ch(n)} \p_{c} (\x) & n \text{~is~a~product~node}, \\
            \sum_{c \in \ch(n)} \! \theta_{n,c} \!\cdot\! \p_{c} (\x)\!\!\!\!\! & n \text{~is~a~sum~node},
        \end{cases}
        \label{eq:pc}
    \end{align}}

\noindent where $f_{n} (\x)$ is an univariate input distribution (\eg Gaussian, Categorical), and $\theta_{n,c}$ denotes the parameter corresponding to edge $(n,c)$. Intuitively, sum nodes model mixtures of their input distributions, which require the mixture weights to be in the probability simplex: $\sum_{c \in \ch(n)} \theta_{n,c} \!=\! 1$ and $\forall c \!\in\! \ch(n), \theta_{n,c} \!\geq\! 0$. And product nodes build factorized distributions over their inputs. The size of a PC, denoted $\abs{\p}$, is the number of edges in its DAG.
\end{defn}

The key to guaranteeing exact and efficient computation of various probabilistic queries is to impose proper structural constraints on the DAG of the PC. As an example, with smoothness and decomposability \citep{poon2011sum}, computing any marginal probability amounts to a forward pass (children before parents) following \cref{eq:pc}, with the only exception that we set the value of input nodes defined on marginalized variables to be $1$. Please refer to \citet{choi20probabilistic} for a comprehensive overview of different structural constraints and what queries they enable.

Although different algorithms are used for different training and inference tasks, they are mostly based on (variants of) the following subroutines: a feedforward pass (Eq.~(\ref{eq:pc})) that computes $\log \p_{n_{\mathrm{r}}} (\x)$, and a backward pass computing 
    {\setlength{\abovedisplayskip}{0.4em}
    \setlength{\belowdisplayskip}{-0.2em}
    \begin{align}
        \forall n, \frac{\partial \log \p_{n_{\mathrm{r}}} (\x)}{\partial \log \p_{n} (\x)} \text{~and~} \forall \theta_{n,c}, \frac{\partial \log \p_{n_{\mathrm{r}}} (\x)}{\partial \theta_{n,c}}.
        \label{eq:pc-gradients}
    \end{align}}

For example, \citet{peharz2020einsum} demonstrate how the above parameter gradients can be used to apply Expectation-Maximization (EM) updates, and \citet{vergari2021compositional} elaborates how the forward pass can be used to compute various probabilistic and information-theoretic queries when coupled with PC structure transformation algorithms. Therefore, the speed and memory efficiency of these two procedures largely determine the overall efficiency of PCs.

% \guy{should we say this is already on the GPU? Before that there was Daniel Lowd's https://libra.cs.uoregon.edu/ which was on CPU}

\boldparagraph{Related work on accelerating PCs.}
There has been a great amount of effort put into speeding up training and inference for PCs. One of the initial attempts performs node-based computations on both CPUs \citep{lowd&rooshenas2015} and GPUs \citep{pronobis2017libspn,molina2019spflow}, \ie by computing the outputs for a mini-batch of inputs (data) recursively for every node. Despite its simplicity, it fails to fully exploit the parallel computation capability possessed by modern GPUs since it can only parallelize over a batch of samples. This problem is mitigated by also parallelizing topologically independent nodes \citep{peharz2020einsum,dang2021juice}. Specifically, a PC is chunked into topological layers, where nodes in the same layer can be computed in parallel. This leads to $1$-$2$ orders of magnitude speedup compared to node-based computation.

The regularity of edge connection patterns is another key factor influencing the design choices. Specifically, EiNets \citep{peharz2020einsum} leverage off-the-shelf Einsum operations to parallelize dense PCs where every layer contains groups of densely connected sum and product/input nodes. \citet{mari2023unifying} generalize the notion of dense PCs to tensorized PCs, which greatly expands the scope of EiNets. \citet{dang2021juice} instead focus on speeding up sparse PCs, where different nodes could have drastically different numbers of edges. They use custom CUDA kernels to balance the workload of different GPU threads and achieve decent speedup on both sparse and dense PCs.

% Due to the sparse nature of PCs, nodes in a layer could have different numbers of edges. This causes workload imbalance problems when we evenly distribute nodes into all processors (\eg threads/blocks in GPUs). To overcome this problem, \citet{dang2021juice} proposes parallelizing over edges within a layer. As illustrated in \cref{fig:edge-parallel-problems}(a), each of the four processors (\eg threads) only needs to process $3$ edges, despite the 2nd and 3rd sum node having $4$ edges.

% Instead of aiming for edge parallelization in the general case, EiNets \citep{peharz2020einsum} focus directly on dense PCs whose sum nodes have the same number of children. \citet{mari2023unifying} further unifies existing dense PCs from the perspective of low-rank tensor decomposition.

Another thread of work focuses on designing computation hardware that is more suitable for PCs. Specifically, \citet{shah2021dpu} propose DAG Processing Units (DPUs) that can efficiently traverse sparse PCs, \citet{dadu2019towards} introduce an indirect read reorder-buffer to improve the efficiency of data-dependent memory accesses in PCs, and \citet{yao2023logarithm} use addition-as-int multiplications to significantly improve the energy efficiency of PC inference algorithms.
% \guy{also https://drive.google.com/file/d/1lI3EfPOVwnjWBiRUG4Pu8zPcCguPwa59/view ?}

\begin{figure}
    \centering
    \includegraphics[width=1.0\columnwidth]{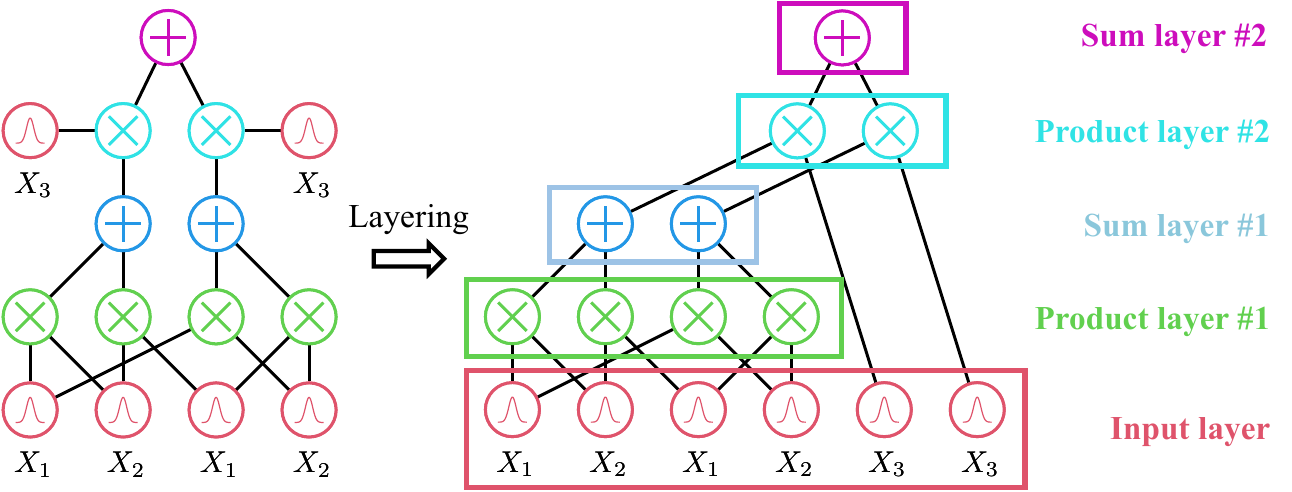}
    \vspace{-2.2em}
    \caption{Layering a PC by grouping nodes with the same topological depth (as indicated by the colors) into disjoint subsets. Both the forward and the backward computation can be carried out independently on nodes within the same layer.}
    \label{fig:layering}
    \vspace{-1.2em}
\end{figure}

\boldparagraph{Applications of PCs.} PCs have been applied to many domains such as explainability and causality \citep{correia2020joints,wang2023compositional}, graph link prediction \citep{loconte2023turn}, lossless data compression \citep{liu2021lossless}, neuro-symbolic AI \citep{xu2018sl, manhaeve18deepproblog,ahmed2022SPLs, ahmed2022entropy}, gradient estimation \citep{ahmed2023simple}, graph neural networks rewiring \citep{qian2023rewiring}, and even large language model detoxification \citep{ahmed2023pseudosl}.

% \vspace{-0.4em}
\section{Key Bottlenecks in PC Parallelization}
\label{sec:problems}
% \vspace{-0.2em}

This section aims to lay out the key bottlenecks to efficient PC implementations. For ease of illustration, we focus solely on the forward pass, and leave the unique challenges posed by the backward pass and their solution to \cref{sec:eff-backward}.

We start by illustrating the layering procedure deployed for PCs. Starting from the input nodes, we perform a topological sort of all nodes, clustering nodes with the same topological depth into a layer. For example, in \cref{fig:layering}, the PC on the left side is transformed into an equivalent layered representation on the right, where nodes of the same color belong to the same layer. The forward pass proceeds by sequentially processing each layer, and finally returns the root node's output. To avoid underflow, all probabilities are stored in the logarithm space. Therefore, product layers just need to sum up the corresponding input log-probabilities, while sum layers compute weighted sums of input log-probabilities utilizing the logsumexp trick.

Assume for now that all nodes in every layer have the same number of children. A straightforward strategy is to parallelize over every node and every sample. Specifically, given a layer of size $M$ and batch size $B$, we need to compute in total $M \!\times\! B$ output values, which are evenly distributed to all processors (\eg thread-blocks in GPUs). We apply this idea to a PC with the PD structure \citep{poon2011sum}. The PC has $\sim\!\!1$M nodes and $\sim\!\!150$M edges. Additionally, all nodes within a layer have the same number of children, making it an ideal testbed for the aforementioned algorithm.

\begin{figure}
    \centering
    \includegraphics[width=\columnwidth]{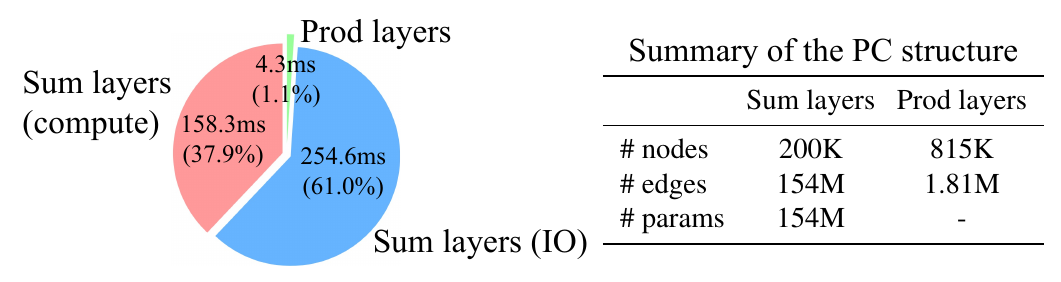}
    \vspace{-2.6em}
    \caption{Runtime breakdown of the feedforward pass of a PC with $\sim\!\!150$M edges. Both the IO and the computation overhead of the sum layers are significantly larger than the total runtime of product layers. Detailed configurations of the PC are shown in the table.}
    % \vspace{-1.4em}
    \label{fig:runtime-breakdown}
\end{figure}

\cref{fig:runtime-breakdown} illustrates the runtime breakdown of the forward pass (with batch size $512$). As shown in the pie chart, both the IO and the computation overhead of the sum layers are much larger than that of the product layers. We would expect sum layers to exhibit a higher computation overhead due to (i) the number of sum edges being $\sim\!\!85$x more than the product edges (see the table in Fig.~\ref{fig:runtime-breakdown}), and (ii) sum edges requiring more compute compared to product edges. However, we would not expect the gap in IO overhead to be as pronounced as indicated in the pie chart. Specifically, with batch size $512$, the ideal memory read count of product layers should be roughly $[\text{batch~size}] \!\times\! [\# \text{sum~nodes}] \!\approx\! 102\text{M}$ since all children of product nodes are sum or input nodes (the number of input nodes is an order of magnitude smaller and is omitted). Similarly, the number of memory reads required by the sum layers is approximately $[\text{batch~size}] \!\times\! [\# \text{prod~nodes}] \!+\! [\# \text{parameters}] \!\approx\! 571\text{M}$, which is only $5.6$x compared to the product layers. The ideal memory write count of product layers should be larger since there are about $4$x more product nodes compared to sum nodes.

While the ideal IO overhead of the sum layers is not much larger than that of the product layers, the drastic difference in runtime (over $50$x) can be explained by the significant amount of reloads of child nodes' probabilities in the sum layers. Specifically, in the adopted PD structure, every sum node has no more than $12$ parents, while most product nodes have $256$ parents.\footnote{Only the children of the root sum node have $1$ parent.} Recall that the parents of product nodes are sum nodes and vice versa. As a result, each sum layer needs to reload the output of every product node multiple times. Although this does not lead to $256$x loads from the GPU's High-Bandwidth Memory (HBM) thanks to its caching mechanism, such excessive IO access still significantly slows down the algorithm.

The fundamental principle guiding our design is to \emph{properly group, or allocate, sum edges to different processors to minimize the reloading of product nodes' outputs}. As an added benefit, this allows us to interpret part of the core computation as matrix multiplications, allowing us to harness Tensor Cores available in modern GPUs and resulting in a significant reduction in sum layers' computational overhead.

\section{Harnessing Block-Based PC Parallelization}
\label{sec:blk-sp-parallel}
% \vspace{-0.2em}

This section takes gradual steps toward demonstrating how we can reduce both the IO and computation overhead using block-based parallelization. Specifically, we first utilize a fully connected sum layer to sketch the high-level idea (Sec.~\ref{sec:fully-connected}). Consequently, we move on to the general case, providing further details of the algorithm (Secs.~\ref{sec:general-case},~\ref{sec:kernel}).

% \vspace{-0.4em}
\subsection{Fully Connected Sum Layers}
\label{sec:fully-connected}
% \vspace{-0.2em}

\begin{figure}
    \centering
    \includegraphics[width=\columnwidth]{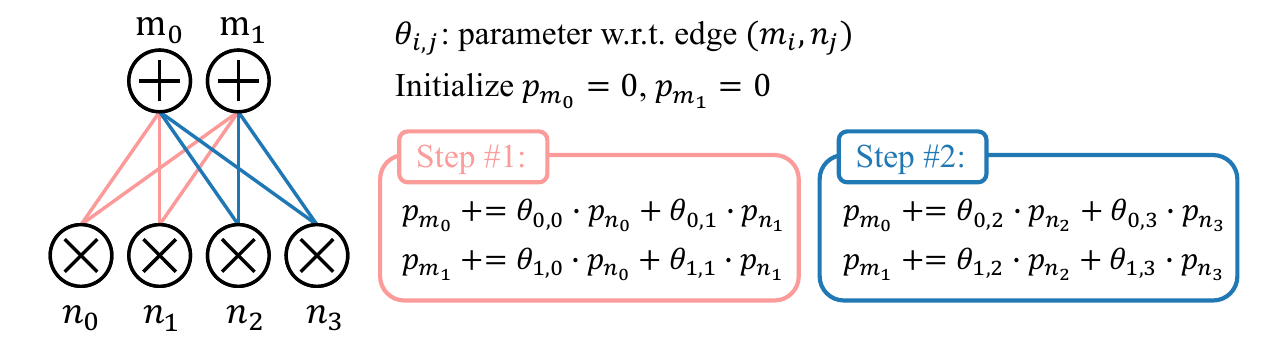}
    \vspace{-2.4em}
    \caption{Illustration of block-based parallelization. A processor computes the output of $2$ sum nodes, by iterating through blocks of $2$ input product nodes and accumulating partial results.}
    \vspace{-1.0em}
    \label{fig:block-parallel-example}
\end{figure}

Consider a fully connected sum layer comprised of $M$ sum nodes, each connected to the same set of $N$ product nodes as inputs. Under the parallelization strategy mentioned in \cref{sec:problems}, with a single sample, we have $M$ processors each computing the output of a sum node. Since the layer is fully connected, every processor loads all $N$ input log-probabilities, which results in $M$ reloads of every input.

The key to reducing excessive IO overhead is by parallelizing over blocks of nodes/edges. Specifically, we divide the $M$ sum nodes into blocks of $K_{\!M}$ nodes and the $N$ product nodes into blocks of $K_{\!N}$ nodes. We assume without loss of generality that $M$ and $N$ are divisible by $K_{\!M}$ and $K_{\!N}$, respectively.\footnote{When the number of product and sum nodes are not divisible by the respective block size, we can add at most $K_M - 1$ (or $K_N - 1$) placeholder nodes to make them divisible by the block size. The incurred additional computation overhead can be small since we can achieve good efficiency with relatively small block sizes (e.g., 32 or 64) given that the number of nodes in a layer is typically greater than a few thousand.} Instead of independently computing the output of every sum node, we calculate the $K_{\!M}$ outputs of a sum node block in a single processor. To achieve this, we iterate through every product node block to compute and accumulate the partial results from the $K_{\!M} \!\times\! K_{\!N}$ edges between the corresponding sum node block and product node block. 

In every step, the processor loads a block of $\boldsymbol{\theta} \!\in\! \mathbb{R}^{K_{\!M} \!\times\! K_{\!N}}$ parameters and a vector of $\boldsymbol{\p}_{\mathrm{prod}} \!\in\! \mathbb{R}^{K_{\!N}}$ input probabilities, where we (temporarily) omit the fact that all probabilities are stored in the logarithm space. The partial outputs $\boldsymbol{\p}_{\mathrm{sum}} \!\in\! \mathbb{R}^{K_{\!M}}$ are computed via a matrix-vector multiplication between $\boldsymbol{\theta}$ and $\boldsymbol{\p}_{\mathrm{prod}}$. Note that if we add a second ``batch'' dimension to $\boldsymbol{\p}_{\mathrm{prod}}$ and $\boldsymbol{\p}_{\mathrm{sum}}$, the computation immediately becomes a matrix-matrix multiplication, which can be computed efficiently using GPU Tensor Cores. 

For example, in \cref{fig:block-parallel-example}, define $K_{\!M} \!=\! K_{\!N} \!=\! 2$, we compute the output of $m_0$ and $m_1$ by first calculating the weighted sum \wrt the input probability of $n_0$ and $n_1$ in step \#1, and then accumulate the probabilities coming from $n_2$ and $n_3$ in step \#2. With the new parallelization strategy, every processor that computes $K_{\!M}$ output values needs to load every input probability only once, and the number of reloads is reduced from $M$ to $M / K_{\!M}$.

% \vspace{-0.4em}
\subsection{Generalizing To Practical Sum Layers}
\label{sec:general-case}
% \vspace{-0.2em}

% \guy{This subsection is 1.5 page long; is there any way to split it up?}

\begin{figure}[t]
    \centering
    \includegraphics[width=\columnwidth]{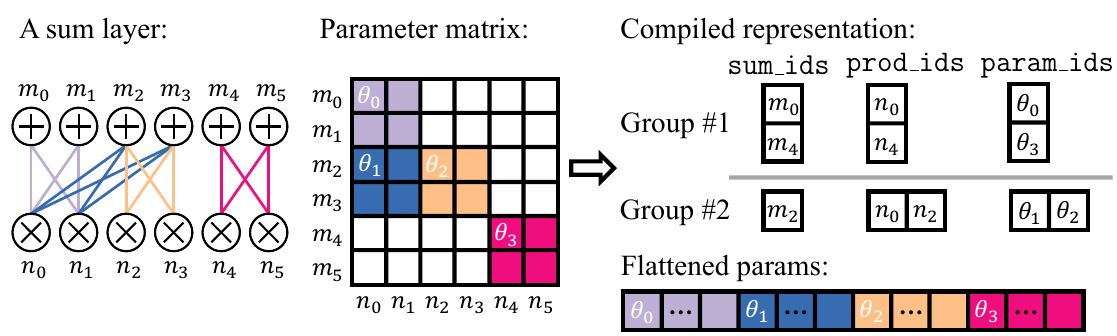}
    \vspace{-2.4em}
    \caption{A sum layer (left) with a block-sparse parameter matrix (middle) is compiled into two kernels (right) each with a balanced workload. During execution, each kernel uses the compiled sum/prod/param indices to compute the outputs of $m_0, \dots, m_5$.}
    \vspace{-1.2em}
    \label{fig:compile-layer}
\end{figure}

Many sum layers in practical PCs are not fully connected (\eg in \citet{dang2022sparse,liu2022scaling}). However, as we shall demonstrate, they can still harness the advantages of block-based parallelization. Specifically, consider a sum layer with $M$ sum nodes and $N$ product nodes as inputs. Following \cref{sec:fully-connected}, we partition the sum and the product nodes into blocks of $K_{\!M}$ and $K_{\!N}$ nodes, respectively. For every pair of sum and product node blocks, if it is either fully connected (\ie featuring $K_{\!M} \!\times\! K_{\!N}$ edges) or unconnected (\ie no edge between them), we call the layer block-sparse. In the following, we focus on efficiently parallelizing block-sparse PCs (whose sum layers all exhibit block-sparsity). We show in \cref{sec:pc-block-sparsity} that many widely-adopted PCs are indeed block sparse w.r.t.\ large block sizes. In \cref{sec:analysis}, we describe how our implementation can speed up sparse PCs. We also show in \cref{sec:exp-faster-pcs} that PyJuice speeds up sparse PCs.

As an example, the layer illustrated in \cref{fig:compile-layer} (left) exhibits block sparsity with block sizes $K_{\!M} \!=\! K_{\!N} \!=\! 2$. This is evident as each pair of sum and product node blocks is either fully connected (\eg $\{m_2, m_3\}$ and $\{n_0, n_1\}$) or disjoint (\eg $\{m_4, m_5\}$ and $\{n_2, n_3\}$). In \cref{fig:compile-layer} (middle), this pattern is more discernible in the parameter matrix, where \emph{aligned} $2 \!\times\! 2$ blocks display either all non-zero parameters (indicated by the colors) or all zero parameters.

Similar to the procedure outlined in \cref{sec:fully-connected}, computing the outputs of a block of $K_{\!M}$ sum nodes involves iterating through all its connected product node blocks. This introduces two additional problems: (i) how to efficiently index the set of connected product node blocks, which may vary for each sum node block; (ii) different sum node blocks could connect to different numbers of product node blocks, which causes an imbalanced workload among processors. For instance, consider the layer in \cref{fig:compile-layer}. The first issue is exemplified by the two sum node blocks $\{m_0, m_1\}$ and $\{m_4, m_5\}$, both of which possess a single child node block, albeit different ones. The second issue is illustrated by the node block $\{m_2, m_3\}$, which connects to two child node blocks, while the others connect to only one.

% \vspace{-0.4em}
\subsection{Efficient Implementations by Compiling PC Layers}
\label{sec:kernel}
% \vspace{-0.2em}

We address both problems through a compilation process, where we assign every node an index, and precompute index tensors that enable efficient block-based parallelization. The first step is to partition the sum node blocks into groups, such that every node block within a group has a similar number of connected child node blocks. We then pad the children with pseudo-product node blocks with probability $0$ such that all sum node blocks in a group have the same number of children. The partition is generated by a dynamic programming algorithm that aims to divide the layer into the smallest possible number of groups while ensuring that the fraction of added pseudo-node blocks does not exceed a pre-defined threshold. Due to space constraints, we elaborate the node block partitioning algorithm in \cref{sec:layer-partitioning}. We also discuss its optimality and time/memory efficiency.

We move on to construct the index tensors for each group. In addition to assigning every node an index, we create a vector $\boldsymbol{\theta}_{\mathrm{flat}}$, a concatentation of all the PC parameters. For every sum node block in a group with $C_{\!N}$ child node blocks, we record (i) the starting index of the sum node block, (ii) the set of initial indices of its $C_{\!N}$ child node blocks, and (iii) the corresponding set of $C_{\!N}$ parameter indices (that point to the first parameter in the respective block of parameters in $\params_{\mathrm{flat}}$). These parameter indices each denote the starting point for the $K_{\!M} \!\times\! K_{\!N}$ parameters of the corresponding pair of sum and product node blocks. Let $C_{\!M}$ represent the total number of node blocks in the group. Following the indices described above, we record the following tensors: $\mathtt{sum\_ids} \!\in\! \mathbb{Z}^{C_{\!M}}$ containing indices of all sum node blocks; $\mathtt{prod\_ids}, \mathtt{param\_ids} \!\in\! \mathbb{Z}^{C_{\!M} \times C_{\!N}}$, whose $i$th row represent the child indices and parameter indices of the $i$th sum node block (\ie the node block with the start index $\mathtt{sum\_ids}[i]$),~respectively.

\begin{figure}[t]
\vspace{-0.8em}
\begin{algorithm}[H]
\caption{Forward pass of a sum layer group}
\label{alg:fwd-pass}
{\fontsize{9}{9} \selectfont
\begin{algorithmic}[1]

\STATE {\bfseries Inputs:} log-probs of product nodes $\boldsymbol{l}_{\mathrm{prod}}$, flattened parameter vector $\boldsymbol{\theta}_{\mathrm{flat}}$, $\mathtt{sum\_ids}$, $\mathtt{prod\_ids}, \mathtt{param\_ids}$

\vspace{0.2em}

\STATE {\bfseries Inputs:} \# sum nodes: $M$, \# product nodes: $N$, batch size: $B$

\vspace{0.2em}

\STATE {\bfseries Inputs:} block sizes $K_{\!M}$, $K_{\!N}$, $K_{\!B}$ for the sum node, product node, and batch dimensions, respectively

\vspace{0.2em}

\STATE {\bfseries Inputs:} number of sum node blocks $C_{\!M}$; number of product node blocks $C_{\!N}$; number of batch blocks $C_{\!B}$

\vspace{0.2em}

\STATE {\bfseries Outputs:} log-probs of sum nodes $\boldsymbol{l}_{\mathrm{sum}}$

\vspace{0.2em}

\STATE {\bfseries Kernel launch:} schedule to launch $C_{\!M} \times C_{\!B}$ thread-blocks with $\mathtt{m} \!=\! 0, \dots, C_{\!M} \!-\! 1$ and $\mathtt{b} \!=\! 0, \dots, C_{\!B} \!-\! 1$

\vspace{0.2em}

\STATE $\mathtt{cum} \leftarrow (-\infty)_{K_{\!M} \!\times\! K_{\!B}} \!\!\in \mathbb{R}^{K_{\!M} \times K_{\!B}}$ \hfill \textcolor[RGB]{115,119,123}{$\triangleright$ Scratch space on SRAM}

\vspace{0.2em}

\STATE $\mathtt{bs}, \mathtt{be} \leftarrow \mathtt{b} \cdot K_{\!B}, (\mathtt{b} + 1) \cdot K_{\!B}$ \hfill \textcolor[RGB]{115,119,123}{$\triangleright$ Start and end batch index}

\vspace{0.2em}

\FOR{\tikzmarknode{a1}{} $\! \mathtt{n} = 0$ \textbf{to} $C_{\!N} \!-\! 1$}

\vspace{0.2em}

\STATE $\mathtt{ps}, \mathtt{ns} \leftarrow \mathtt{param\_ids}[\mathtt{m},\mathtt{n}], \mathtt{prod\_ids}[\mathtt{n},\mathtt{b}]$

\vspace{0.2em}

\STATE Load $\boldsymbol{\theta} \!\leftarrow\! \boldsymbol{\theta}_{\mathrm{flat}}[\mathtt{ps} \!:\! \mathtt{ps} + K_{\!M} \!\cdot\! K_{\!N}].\mathtt{view}(\!K_{\!M}, \!K_{\!N}\!)$ to SRAM

\vspace{0.2em}

\STATE Load $\boldsymbol{l} \!\leftarrow\! 
\boldsymbol{l}_{\mathrm{prod}}[\mathtt{ns} \!:\! \mathtt{ns} + K_{\!N},\mathtt{bs} \!:\! \mathtt{be}] \!\in\! \mathbb{R}^{K_{\!N} \times K_{\!B}}$ to SRAM

\vspace{0.2em}

\STATE $\boldsymbol{l}_{\mathrm{max}} \leftarrow \mathtt{max}(\boldsymbol{l}, \mathtt{dim}\!=\!0) \in \mathbb{R}^{1 \times K_{\!B}}$ \hfill \textcolor[RGB]{115,119,123}{$\triangleright$ Compute on chip}

\vspace{0.2em}

\STATE $\boldsymbol{\p}_{\mathrm{p}} \leftarrow \mathtt{exp}(\boldsymbol{l} - \boldsymbol{l}_{\mathrm{max}}) \in \mathbb{R}^{K_{\!N} \times K_{\!B}}$

\vspace{0.2em}

\STATE $\boldsymbol{\p}_{\mathrm{s}} \leftarrow \mathtt{matmul}(\boldsymbol{\theta}, \boldsymbol{\p}_{\mathrm{p}}) \in \mathbb{R}^{K_{\!M} \times K_{\!B}}$ \hfill \textcolor[RGB]{115,119,123}{$\triangleright$ With Tensor Cores}

\vspace{0.3em}

\STATE \hspace{-0.5em}
    {\setlength{\abovedisplayskip}{-1.0em}
    \setlength{\belowdisplayskip}{-0.2em}
    \begin{align*}
    \fontsize{9}{9}\selectfont
    \!\!\!\!\mathtt{cum} \leftarrow \mathtt{where}( & \boldsymbol{l}_{\mathrm{max}} > \mathtt{cum}, \\[-0.1em]
    & \mathtt{log}(\boldsymbol{\p}_{\mathrm{s}} + \mathtt{exp}(\mathtt{cum} - \boldsymbol{l}_{\mathrm{max}}) + \boldsymbol{l}_{\mathrm{max}}, \\[-0.1em]
    & \mathtt{log}(\mathtt{exp}(\boldsymbol{l}_{\mathrm{max}} - \mathtt{cum}) \cdot \boldsymbol{\p}_{\mathrm{s}} + 1) + \mathtt{cum})
\end{align*}}

\ENDFOR

\vspace{0.3em}

\STATE $\boldsymbol{l}_{\mathrm{sum}}[\mathtt{ms} \!:\! \mathtt{ms} + K_{\!M},\mathtt{bs} \!:\! \mathtt{be}] \!\leftarrow\! \mathtt{acc} \;\;$ (where $\mathtt{ms} \!\leftarrow\! \mathtt{sum\_ids}[\mathtt{m}]$)
    
\end{algorithmic}
}    
\end{algorithm}
\begin{tikzpicture}[overlay,remember picture]
    \draw[black,line width=0.6pt] ([xshift=-12pt,yshift=-3pt]a1.west) -- ([xshift=-12pt,yshift=-110pt]a1.west) -- ([xshift=-8pt,yshift=-110pt]a1.west);
\end{tikzpicture}
\vspace{-3.0em}
\end{figure}

\cref{fig:compile-layer} (right) illustrates the compiled index tensors of the sum layer shown on the left. Recall that we use the block sizes $K_{\!M} \!=\! K_{\!N} \!=\! 2$. The layer is then divided into two groups: the first group including two sum node blocks, $\{m_0, m_1\}$ and $\{m_4, m_5\}$, each having one child node block, and the second group including one sum node block, $\{m_2, m_3\}$, which has two child node blocks. Take, for instance, the first group. $\mathtt{sum\_ids}$ stores the start indices (\ie $m_0$ and $m_4$) of the two sum node blocks. $\mathtt{prod\_ids}$ stores the initial indices of the child node blocks (\ie $n_0$ and $n_4$) of the two sum node blocks, respectively. $\mathtt{param\_ids}$ encodes the corresponding initial parameter indices $\theta_0$ and~ $\theta_2$.

Partitioning a layer into groups with the same number of children allows us to use different kernel launching hyperparameters according to the specific setup of every node group (\eg number of nodes) to achieve better performance.

% \guy{I think we need to state more directly why it helps that within one kernel invocation, all blocks have the same number of inputs. It helps achieve different parallelization strategies for different groups, where the node/edge tradeoffs are different and the thread blocks need to be allocated differently in these dimensions. Something like this?}
For every group in a sum layer, the three index tensors serve as inputs to a CUDA kernel computing the log-probabilities of the sum nodes in the group. Define $\boldsymbol{l}_{\mathrm{prod}} \!\in\! \mathbb{R}^{N \times B}$ and $\boldsymbol{l}_{\mathrm{sum}} \!\in\! \mathbb{R}^{M \times B}$ ($B$ is the batch size) as the set of input and output log-probabilities, respectively. Consider a group with $C_{\!M}$ sum node blocks and $C_{\!N}$ child node blocks per sum node block. \cref{alg:fwd-pass} computes the log-probabilities of the $C_{\!M}$ sum node blocks and stores the results in the proper locations in $\boldsymbol{l}_{\mathrm{sum}}$. Specifically, we also divide the $B$ samples into blocks of size $K_{\!B}$, leading to $C_{\!B} \!:=\! B / K_{\!B}$ blocks (assume w.l.o.g. that $B$ is divisible by $K_{\!B}$). \cref{alg:fwd-pass} schedules to launch $C_{\!M} \!\times\! C_{\!B}$ thread-blocks, each responsible for computing $K_{\!M} \!\times\! K_{\!B}$ outputs (line 6). The main loop in line 9 iterates over all $C_{\!N}$ child node blocks. In every step, we first load the corresponding parameter matrix $\boldsymbol{\theta} \!\in\! \mathbb{R}^{K_{\!M} \times K_{\!N}}$ (line 11) and input matrix $\boldsymbol{l} \!\in\! \mathbb{R}^{K_{\!N} \times K_{\!B}}$ (line 12). Since $\boldsymbol{l}$ contains log-probabilities, we apply a variant of the logsumexp trick: we first convert $\boldsymbol{l}$ to the arithmetic space by subtracting the per-sample maximum log-probability (lines 13-14), then compute the (partial) output probabilities from the current set of $K_{\!M} \!\times\! K_{\!N}$ edges via matrix multiplication (line 15), and in line 16 aggregate the results back to the accumulator $\mathtt{cum}$ defined in line 7. Finally, we store the log-probabilities to the target locations in $\boldsymbol{l}_{\mathrm{sum}}$ (line 17).

\begin{figure}[t]
    \centering
    \includegraphics[width=1.0\columnwidth]{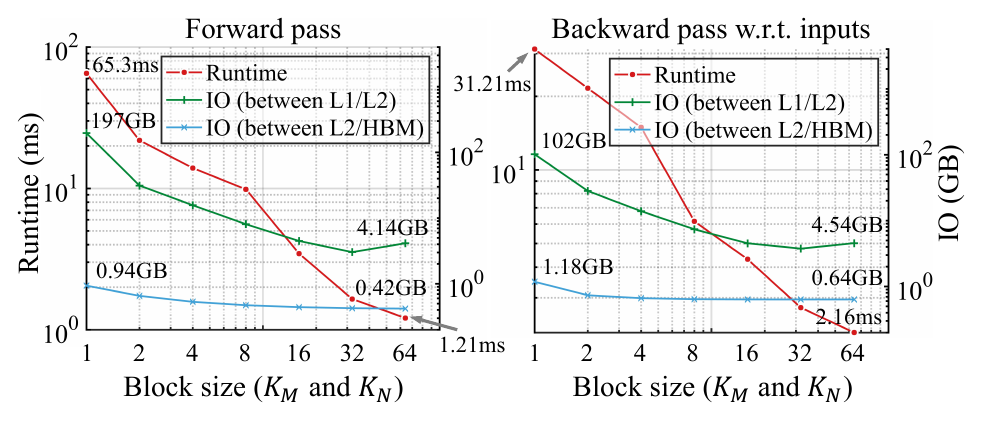}
    \vspace{-2.4em}
    \caption{Runtime and IO overhead of a sum layer from the PD structure (with $29$K nodes and $30$M edges). The results demonstrate significant performance gains from our block-based parallelization, even with small block sizes.}
    \label{fig:analysis}
    % \vspace{-1.2em}
\end{figure}

\subsection{Analysis: IO and Computation Overhead}
\label{sec:analysis}

We analyze the efficiency and IO complexity of our block-based parallelization strategy. Specifically, we benchmark on the largest sum layer in the PD structure adopted in \cref{sec:problems}. The layer consists of $29$K nodes and $30$M edges. In addition to the computation time, we record two types of IO overhead: (i) the IO between the L1/texture cache and the L2 cache, and (ii) the reads/writes between the L2 cache and the GPU High-Bandwidth Memory (HBM). We vary the block sizes $K_{\!M}$ and $K_{\!N}$ exponentially from $1$ to $64$. To ensure a fair comparison, we implement a dedicated kernel for $K_{\!M} \!=\! K_{\!N} \!=\! 1$, which directly parallelizes over sum node/sample pairs, allowing for better workload allocation. For other block sizes, we adjust $K_{\!B}$ and other kernel launching hyperparameters (\eg warps per block) and report the best runtime for every case. Results of the backward pass (\wrt inputs) are also reported for completeness.

Results are shown in \cref{fig:analysis}. As the block size increases, both the forward and the backward pass become significantly faster. Notably, this is accompanied by a significant drop in IO overhead. Specifically, with a large block size, the kernel consumes $2$x fewer reads/writes between the L2 cache and the HBM, and $25$-$50$x fewer IO between the L1 and L2 cache. This corroborates the hypothesis stated in \cref{sec:problems} that the extensive value reloads significantly slow down the computation.

% Results are illustrated in \cref{fig:analysis}. We first focus on the runtime. With larger block sizes, both the forward and the backward pass become significantly faster. A seemingly counter-intuitive observation is that the backward pass (\wrt inputs) is faster than the forward pass when the block size is small, and is surpassed by the forward pass when a larger block size is used. This is attributed to our observation that the backward pass can be computed directly on the arithmetic space without suffering from numerical stability problems, which saves a considerable amount of log/exp operations. When the block size is large, the speedup gained by this trick is out-paced by the IO overhead of additional values to be loaded (\eg ``gradients'' of the sum nodes).

% Next, with large block size, the kernel consumes $2$x fewer reads/writes between the L2 cache and HBM, and $25-50$x fewer IO between the L1 and L2 cache. This corroborates the hypothesis stated in \cref{sec:problems} that the extensive value reloads of edge parallelization slow down the kernel.

Additionally, we note that even with small block sizes (\eg $2$ or $4$), the speedup is quite significant compared to the baseline case ($K_{\!M} \!=\! K_{\!N} \!=\! 1$), which allows us to speed up \emph{sparse} PCs. Specifically, with the observation that every sparse PC can be viewed as a block-sparse PC with block size $1$, we can transform a sparse PC into a block-sparse one, and pad zero parameters to edges belonging to the block-sparse PC but not the sparse PC. For PCs with relatively regular sparsity patterns, increasing the block sizes to even small values like 2 or 4 can lead to significant speedup even though a relatively large number of pseudo edges need to be padded.

the speedup obtained by having a larger block size outpaces the overhead caused by padded edges with zero parameters, which leads to speed-ups.

% \vspace{-0.4em}
\section{Optimizing Backpropagation with PC Flows}
\label{sec:eff-backward}
% \vspace{-0.2em}

The previous section focuses on speeding up sum layers by reducing excessive memory reloads and leveraging Tensor Cores. However, when it comes to backpropagation, directly adapting \cref{alg:fwd-pass} by differentiating lines $13$-$16$ would lead to poor performance due to the following. First, we need to either store some intermediate values (\eg $\boldsymbol{l}_{\mathrm{max}}$ and $\boldsymbol{\p}_{\mathrm{p}}$) in the forward pass or recompute them in the backward pass. Next, since different thread-blocks could access the same product node log-probabilities in line 12, they both need to write (partial) gradients of it, which introduces inter-thread-block barriers that slow down the execution.

We overcome the problems by leveraging PC flows \citep{choi21group}, which is only a factor of $\theta_{n,c}$ away 
% \guy{I am confused about this factor of $\theta_{n,c}$; the equations below do not contain a factor of $\theta_{n,c}$; I guess it is implicit in the log in the denominator of the partial derivative...} 
from the desired gradients (Eq.~\ref{eq:pc-gradients}). PC flows exhibit a straightforward recursive definition, facilitating a seamless transformation into an efficient implementation for the backward pass.

\begin{defn}[PC flows]
\label{def:pc-flows}
For a PC $\p_{n_r} (\X)$ rooted at node $n_r$ and a sample $\x$, the flow $\mathrm{F}_{n}(\x)$ of every node $n$ is defined recursively as follows (assume that no consecutive sum nodes or product nodes exist in the PC):\footnote{If such nodes exist, we can always collapse them into a single sum or product node.}
    {\setlength{\abovedisplayskip}{0.4em}
    \setlength{\belowdisplayskip}{-0.2em}
    \begin{align*}
        \mathrm{F}_{n} (\x) \!:=\! \begin{cases}
            \quad \; 1 & n \text{~is~the~root~node}, \\
            \sum\limits_{m \in \pa(n)} \!\!\!\mathrm{F}_{m} (\x) & n \text{~is~input~or~sum}, \\
            \sum\limits_{m \in \pa(n)} \!\!\!\!\frac{\theta_{m,\!n} \cdot \p_{n} (\x)}{\p_{m} (\x)} \!\cdot\! \mathrm{F}_{m} (\x)\!\!\!\! & n \text{~is~a~product~node},
        \end{cases}
    \end{align*}}

\noindent where $\pa(n)$ is the set of parents of $n$. Similarly, the edge flow $\mathrm{F}_{n,c} (\x)$ \wrt the sample $\x$ ($c \!\in\! \ch(n)$) is defined as
    \begin{align*}
        \mathrm{F}_{n,c} (\x) := \theta_{n,c} \cdot \p_{c} (\x) / \p_{n} (\x) \cdot \mathrm{F}_{n} (\x).
    \end{align*}
\end{defn}

While similar results have been established in a slightly different context \citep{peharz2020einsum}, we prove the following equations in \cref{sec:flow-grad-rel} for completeness:
    {\setlength{\abovedisplayskip}{0.4em}
    \setlength{\belowdisplayskip}{-0.2em}
    \begin{align*}
        \mathrm{F}_{n} (\x) = \frac{\partial \log \p_{n_r} (\x)}{\partial \log \p_{n} (\x)} \text{~and~} \mathrm{F}_{n,c} (\x) = \theta_{n,c} \!\cdot\! \frac{\partial \log \p_{n_r} (\x)}{\partial \theta_{n,c}}.
    \end{align*}}

Following \cref{def:pc-flows}, we can compute $F_{n} (\x)$ for every node $n$ utilizing the same set of layers created for the feedforward pass. Specifically, we first set the flow of the root node to $1$ following its definition. We then iterate through the layers in reverse order (\ie parent layers before child layers). While processing a layer, all flows of the nodes in the layer are computed by the preceding layers. And our goal is to compute the (partial) flows of the child nodes of the layer. Similar to the forward pass, we compile every layer by grouping child node blocks with a similar number of parents, and use block-based parallelization to reduce reloads of parent log-probabilities. We provide the full details of the backpropagation algorithm in \cref{sec:backprop-details}.

Another important design choice that leads to a significant reduction in memory footprint is to recompute the product nodes' probabilities in the backward pass instead of storing them all in the GPU memory during the forward pass. Specifically, we maintain a scratch space on GPU HBM that can hold the results of the largest product layer. All product layers write their outputs to this same scratch space, and the required product node probabilities are re-computed when requested by a sum layer during backpropagation. Since product layers are extremely fast to evaluate compared to the sum layers (\eg see the runtime breakdown in Fig.~\ref{fig:runtime-breakdown}), this leads to significant memory savings at the cost of slightly increased computation time.

% \guy{there is nothing interesting to say about the input layer?}

% \vspace{-0.4em}
\section{Experiments}
\label{sec:exps}
% \vspace{-0.2em}

We evaluate the impact of using PyJuice to train PC models. In \cref{sec:exp-faster-pcs}, we compare PyJuice against existing implementations regarding time and memory efficiency. To demonstrate its generality and flexibility, we evaluate PyJuice on four commonly used dense PC structures as well as highly unstructured and sparse PCs. Next, we demonstrate that PyJuice can be readily used to scale up PCs for various downstream applications in \cref{sec:exp-scaling-up}. Finally, in \cref{sec:exp-benchmarking}, we benchmark existing PCs on high-resolution image datasets, hoping to incentivize future research to develop better PC structures as well as learning algorithms.

% \vspace{-0.4em}
\subsection{Faster Models with PyJuice}
\label{sec:exp-faster-pcs}
% \vspace{-0.2em}

We first benchmark the runtime of PyJuice on four commonly used PC structures: PD \citep{poon2011sum}, RAT-SPN \citep{peharz2020random}, HCLT \citep{liu2021tractable}, and HMM \citep{rabiner1986introduction}. For all models, we record the runtime to process 60,000 samples (including the forward pass, the backward pass, and mini-batch EM updates). We vary their structural hyperparameters and create five PCs for every structure with sizes (\ie number of edges) ranging from $500$K to $2$B. We compare against four baselines: SPFlow \citep{molina2019spflow}, EiNet \citep{peharz2020einsum}, Juice.jl \citep{dang2021juice}, and Dynamax \citep{murphy2023dynamax}. Dynamax is dedicated to State Space Models so it is only used to run HMMs; SPFlow and EiNet are excluded in the HMM results because we are unable to construct homogeneous HMMs with their frameworks due to the need to share the transition and emission parameters at different time steps. We describe how PyJuice implements PCs with tied parameters in \cref{sec:param-tying}. All experiments in this subsection are carried out on an RTX 4090 GPU with 24GB memory.

\begin{figure}[t]
    \centering
    % \vspace{-1.2em}
    \includegraphics[width=\columnwidth]{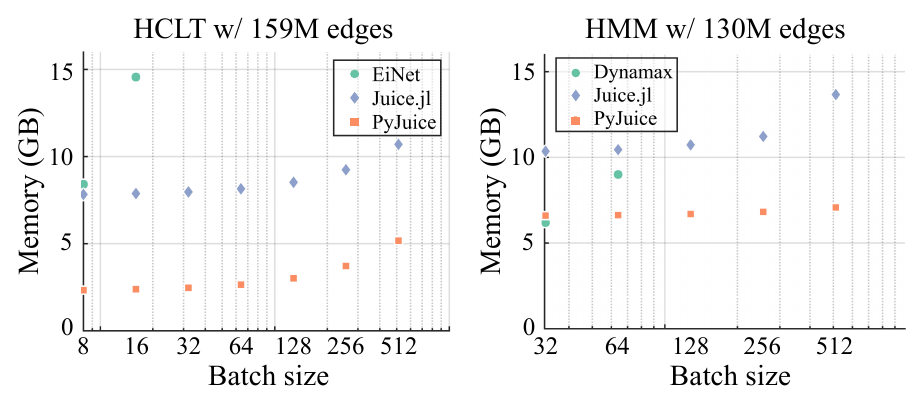}
    \vspace{-3.0em}
    \caption{Comparison on memory efficiency. We take two PCs (\ie an HCLT w/ 159M edges and an HMM w/ 130M edges) and record GPU memory usage under different block sizes.\footnotemark}
    \vspace{-1.0em}
    \label{fig:mem-overhead}
\end{figure}

\footnotetext{In the adopted HMM, running Dynamax with batch size $\geq\!128$ leads to internal errors, and thus the results are not reported.}

\cref{tab:speed-results} reports the runtime in seconds per epoch with mini-batch EMs. PyJuice is orders of magnitude faster than all baselines in both small and large PCs. Further, we observe that most baselines exhaust 24GB of memory for larger PCs (indicated by ``OOM'' in the table), while PyJuice can still efficiently train these models. Additionally, in \cref{sec:exp-compilation-speed}, we show the efficiency of the compilation process. For example, it takes only $\sim\!8.7$s to compile an HCLT with $159$M edges. Note that we only compile the PC once and then reuse the compiled structure for training and inference.

In \cref{fig:mem-overhead}, we take two PCs to show the GPU memory consumption with different batch sizes. The results demonstrate that PyJuice is more memory efficient than the baselines, especially in the case of large batch sizes (note that we always need a constant-size space to store the parameters).

\begin{figure}[H]
    \centering
    \vspace{-0.6em}
    \includegraphics[width=\columnwidth]{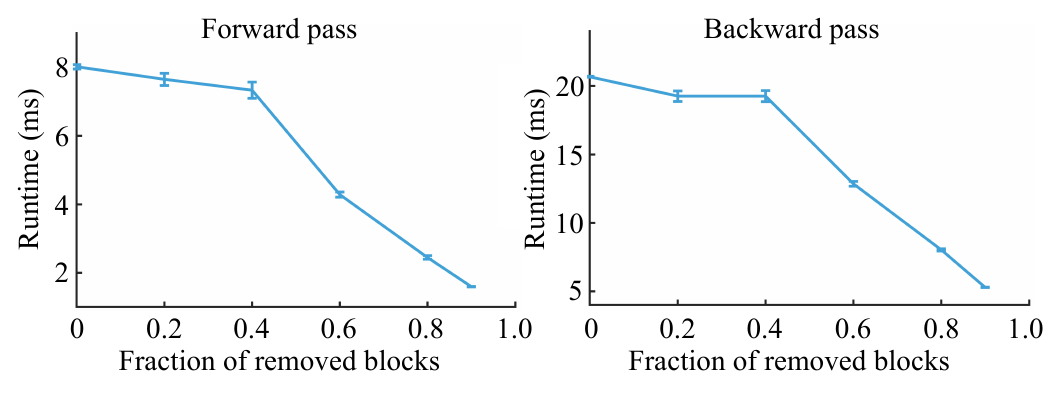}
    \vspace{-2.8em}
    \caption{Runtime of a block-sparse sum layer as the function of the fraction of kept (non-dropped) edge blocks. The error bars represent standard deviations over $5$ runs.}
    \vspace{-1.0em}
    \label{fig:block-sparse-speedup}
\end{figure}

We move on to benchmark PyJuice on block-sparse PCs. We create a sum layer with 209M edges (see Appx.~\ref{sec:exp-blk-sp-layer} for details). We partition the sum and input product nodes in the layer into blocks of $32$ nodes respectively. We randomly discard blocks of $32 \!\times\! 32$ edges, resulting in block-sparse layers. As shown in \cref{fig:block-sparse-speedup}, as the fraction of removed edge blocks increases, the runtime of both the forward and the backward pass decreases significantly. This motivates future work on PC modeling to focus on designing effective block-sparse PCs.

\begin{figure}[H]
    \centering
    \vspace{-1.0em}
    \includegraphics[width=\columnwidth]{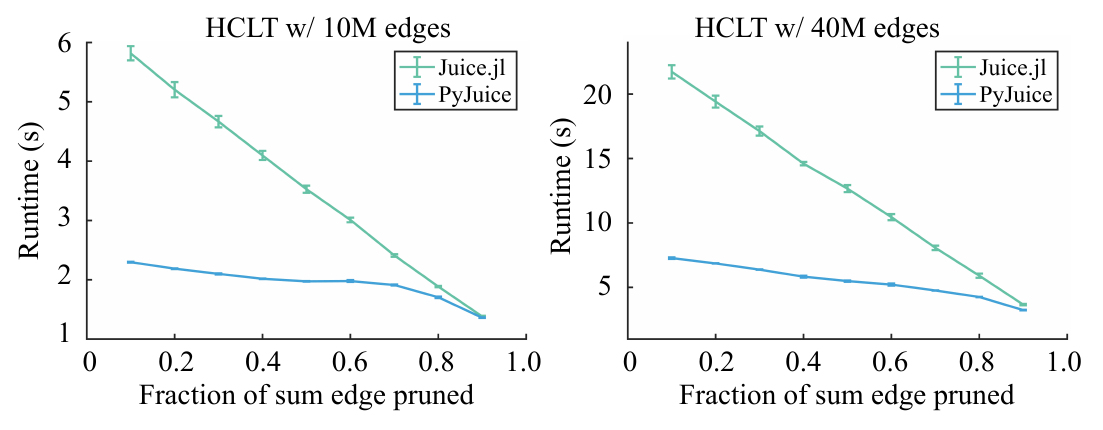}
    \vspace{-2.8em}
    \caption{Runtime per epoch (with 60K samples) of two sparse HCLTs with different fractions of pruned edges. The error bars represent standard deviations over $5$ runs.}
    \vspace{-1.2em}
    \label{fig:pruning-speed}
\end{figure}

Finally, we proceed to evaluate the runtime of sparse PCs. We adopt the PC pruning algorithm proposed by \citet{dang2022sparse} to prune two HCLTs with 10M and 40M edges, respectively. We only compare against Juice.jl since all other implementations do not support sparse PCs. As shown in \cref{fig:pruning-speed}, PyJuice is consistently faster than Juice.jl, despite the diminishing gap when over 90\% edges are pruned. Note that with sparse PCs, PyJuice cannot fully benefit from the block-based parallelization strategy described in \cref{sec:blk-sp-parallel}, yet it can still outperform the baseline. 
% We leave the idea described in the last paragraph of \cref{sec:analysis} to future work.

\subsection{Better PCs At Scale}
\label{sec:exp-scaling-up}
% \vspace{-0.2em}

This section demonstrates the ability of PyJuice to improve the state of the art by simply using larger PCs and training for more epochs thanks to its speed and memory efficiency. Specifically, we take the HMM language model proposed by \citet{zhang2023tractable} and the image model introduced by \citet{liu2023understanding} as two examples.

\boldparagraph{HMM language models.}
\citet{zhang2023tractable} use the Latent Variable Distillation (LVD) \citep{liu2022scaling} technique to train an HMM with 4096 hidden states on sequences of $32$ word tokens. Specifically, LVD is used to obtain a set of ``good'' initial parameters for the HMM from deep generative models. The HMM language model is then fine-tuned on the CommonGen dataset \citep{lin2020commongen}, and is subsequently used to control the generation process of (large) language models for constrained generation tasks. Following the same procedure, we use PyJuice to fine-tune two HMMs with hidden sizes 4096 and 8192, respectively.

\begin{table}[t]
    \centering
    \vspace{-0.4em}
    \caption{Perplexity of HMM language models trained on the CommonGen benchmark \citep{lin2020commongen}.}
    \label{tab:gelato-hmm}
    \vspace{0.2em}
    \renewcommand{\arraystretch}{0.88}
    \centering
    \scalebox{0.84}{
    \begin{tabular}{l@{\hspace{0.4em}}ccc}
        \toprule
         & \citet{zhang2023tractable} & \multicolumn{2}{c}{PyJuice} \\
         \cmidrule(lr){2-2}
         \cmidrule(lr){3-4}
         \# hidden states & 4096 & 4096 & 8192 \\
         \midrule
         Perplexity & 9.78 & 8.81 & \textbf{8.65} \\
        \bottomrule
    \end{tabular}}
% \vspace{-1.4em}
\end{table}

As shown in \cref{tab:gelato-hmm}, by using the same HMM with 4096 hidden states, PyJuice improved the perplexity by $\sim\!1.0$ by running many more epochs in less time compared to the original model. We also train a larger HMM with 8192 hidden states and further improved the perplexity by a further $0.16$. We refer the reader to \cref{sec:exp-gelato-hmm} for more details.

\boldparagraph{Sparse Image Models.}
\citet{liu2023understanding} design a PC learning algorithm that targets image data by separately training two sets of PCs: a set of sparse patch-level PCs (\eg $4 \!\times\! 4$ patches) and a top-level PC that aggregates outputs of the patch-level PC. In the final training step, the PCs are supposed to be assembled and jointly fine-tuned. However, due to the huge memory consumption of the PC (with over 10M nodes), only the top-level model is fine-tuned in the original paper. With PyJuice, we can fit the entire model in 24GB of memory and fine-tune the entire model. For the PC trained on the ImageNet32 dataset \citep{deng2009imagenet}, this fine-tuning step leads to an improvement from $4.06$ to $4.04$ bits-per-dimension. See \cref{sec:exp-lvd-pg} for more details.

% \vspace{-0.4em}
\subsection{Benchmarking Existing PCs}
\label{sec:exp-benchmarking}
% \vspace{-0.2em}

We use PyJuice to benchmark the performance of the PD and the HCLT structure on three natural image datasets: ImageNet \citep{deng2009imagenet} and its down-sampled version ImageNet32, and CelebA-HQ \citep{liu2015faceattributes}. For all three datasets, we train the PCs on randomly sampled $16 \!\times\! 16$ patches\footnote{Note that the top-left position of the patches is sampled uniformly. This produces different results compared to using aligned patches (\eg dividing a 32$\times$32 image into four 16$\times$16 aligned patches).}, which results in a total of $16 \!\times\! 16 \!\times\! 3 \!=\! 768$ categorical variables each with $2^8 \!=\! 256$ possible values. As a preprocessing step, the image patches are converted into the YCoCg color space since it is observed that such color space transformations lead to improved density estimation performance. Note that due to the lossy transformation between the RGB space and the YCoCg space, our results are not directly comparable to the results obtained from RGB images.

We adopt two PD structures (\ie PD-mid with 107M edges and PD-large with 405M edges) as well as two HCLT structures (\ie HCLT-mid with 40M edges and HCLT-large with 174M edges). Details of the adopted models are described in \cref{sec:exp-benchmark-details}. We experiment with different optimization strategies and adopt full-batch EM as it yields consistently better performance across models and datasets. Specifically, the computed PC flows are accumulated across all samples in the training set before doing one EM step.

\begin{table}[t]
    \centering
    % \vspace{-1.2em}
    \caption{Density estimation performance of PCs on three natural image datasets. Reported numbers are test set bits-per-dimension.}
    \label{tab:benchmarking}
    \vspace{0.2em}
    \renewcommand{\arraystretch}{0.88}
    \centering
    \scalebox{0.84}{
    \begin{tabular}{lcccc}
        \toprule
        Dataset & PD-mid & PD-large & HCLT-mid & HCLT-large \\
        \midrule
        ImageNet32 & 5.22 & 5.20 & 4.36 & 4.33 \\
        ImageNet & 4.98 & 4.95 & 3.57 & 3.53 \\
        CelebA-HQ & 4.35 & 4.29 & 2.43 & 2.38 \\
        \bottomrule
    \end{tabular}}
% \vspace{-1.4em}
\end{table}

Results are shown in \cref{tab:benchmarking}. Notably, we achieve \emph{better} results compared to previous papers. For example, \citet{liu2022scaling} reports $4.82$ bits-per-dimension (bpd) for HCLT on ImageNet32, while we achieved $4.33$ bpd. The performance improvements stem from more training epochs and the ability to do more hyperparameter search thanks to the speedup. We highlight that the goal of this section is not to set new records for tractable deep generative models, but to establish a set of baselines that can be easily reproduced to track the progress of developments in PC modeling and learning. In \cref{sec:exp-benchmark-details}, we include additional benchmark results on the WikiText dataset \citep{merity2016pointer}.

% \vspace{-0.4em}
\section{Conclusion}
% \vspace{-0.2em}

We proposed PyJuice, a novel system that supports training and inference of probabilistic circuits. PyJuice is orders of magnitude faster and much more memory efficient than even very recent baselines. We hope PyJuice can boost future research on tractable deep generative models by allowing for efficient training of large-scale~architectures.

\section*{Acknowledgements}

This work was funded in part by the DARPA PTG Program under award HR00112220005, the DARPA ANSR program under award FA8750-23-2-0004, and the NSF grant \#IIS-1943641. We thank Honghua Zhang, Pasha Khosravi, and Poorva Garg for providing valuable feedback during the development of PyJuice.

\section*{Impact Statement}

This paper presents work whose goal is to advance the field of Machine Learning. There are many potential societal consequences of our work, none which we feel must be specifically highlighted here.

\bibliography{refs}
\bibliographystyle{icml2024}

%%%%%%%%%%%%%%%%%%%%%%%%%%%%%%%%%%%%%%%%%%%%%%%%%%%%%%%%%%%%%%%%%%%%%%%%%%%%%%%
%%%%%%%%%%%%%%%%%%%%%%%%%%%%%%%%%%%%%%%%%%%%%%%%%%%%%%%%%%%%%%%%%%%%%%%%%%%%%%%
% APPENDIX
%%%%%%%%%%%%%%%%%%%%%%%%%%%%%%%%%%%%%%%%%%%%%%%%%%%%%%%%%%%%%%%%%%%%%%%%%%%%%%%
%%%%%%%%%%%%%%%%%%%%%%%%%%%%%%%%%%%%%%%%%%%%%%%%%%%%%%%%%%%%%%%%%%%%%%%%%%%%%%%
\newpage
\appendix
\onecolumn

\section{Algorithm Details}

In this section, we provide additional details of the design of PyJuice. Specifically, we introduce the layer partitioning algorithm that divides a layer into groups of node blocks with a similar number of children in \cref{sec:layer-partitioning}, and describe the details of the backpropagation algorithm in \cref{sec:backprop-details}.

\subsection{The Layer Partitioning Algorithm}
\label{sec:layer-partitioning}

The layer partitioning algorithm receives as input a vector of integers $\mathtt{nchs}$ where each number denotes the number of child node blocks connected to a node block in the layer. It also receives as input the maximum number of groups to be considered (denoted $\mathtt{G}$) and a sparsity tolerance threshold $\mathtt{tol} \!\in\! (0, 1]$. Our goal is to search for a set of $n$ (at most $\mathtt{G}$) groups with capacities $g_1, \dots, g_n$, respectively. Every number in $\mathtt{nchs}$ is then placed into the group with the smallest capacity it can fit in. Every number in $\mathtt{nchs}$ must fit in a group. Assume there are $k_i$ numbers assigned to group $i$, the overhead/cost \wrt a partitioning $\{g_1, \dots, g_n\}$ is defined as $\sum_{i \in [n]} k_i \!\cdot\! g_i$. Our goal is to find a partitioning with overhead smaller than $\mathtt{sum}(\mathtt{nchs}) \!\cdot\! (1 \!+\! \mathtt{tol})$.

\begin{figure}[H]
\vspace{-1.8em}
\begin{algorithm}[H]
\caption{Partition a layer into groups}
\label{alg:layer-partitioning}
{\fontsize{9}{9} \selectfont
\begin{algorithmic}[1]

\STATE {\bfseries Inputs:} a list of child node (block) counts of the current layer $\mathtt{nchs} \!\in\! \mathbb{Z}^{N}$ ($N$ is the number of node blocks in the layer)

\vspace{0.2em}

\STATE {\bfseries Inputs:} the maximum number of groups $\mathtt{G}$, the sparsity tolerance threshold $\mathtt{tol} \!\in\! (0,1]$

\vspace{0.2em}

\STATE $\mathtt{uni\_nchs}, \mathtt{counts} \leftarrow \mathtt{unique} (\mathtt{nchs}, \mathtt{sorted}=\mathtt{True})$ (get the unique values and their appearance counts; we require the numbers in $\mathtt{uni\_nchs}$ to be sorted in ascending order)

\vspace{0.2em}

\STATE $\mathtt{L} \leftarrow \mathtt{length} (\mathtt{uni\_nchs})$

\vspace{0.2em}

\STATE $\mathtt{target\_overhead} \leftarrow \lceil \mathtt{sum} (\mathtt{uni\_nchs} * \mathtt{counts}) * (1.0 + \mathtt{tol}) \rceil$ (get the target overhead)

\vspace{0.2em}

\STATE $\mathtt{cum\_counts} \leftarrow \mathtt{cumsum} (\mathtt{counts})$

\vspace{0.2em}

\STATE $\mathtt{dp}, \mathtt{backtrace} \leftarrow (0)_{\mathtt{L} \times \mathtt{G} + 1} \in \mathbb{R}^{\mathtt{L} \times \mathtt{G} + 1}, (0)_{\mathtt{L} \times \mathtt{G} + 1} \in \mathbb{Z}^{\mathtt{L} \times \mathtt{G} + 1}$

\vspace{0.2em}

\FOR{\tikzmarknode{a0}{} $i = 0$ \textbf{to} $\mathtt{L} - 1$}

\STATE $\mathtt{dp}[i,1] \leftarrow \mathtt{uni\_nchs}[i] * \mathtt{cum\_counts}[i]$

\ENDFOR

\vspace{0.2em}

\STATE \textcolor[RGB]{115,119,123}{\# Main DP algorithm}

\vspace{0.2em}

\STATE $\mathtt{target\_n\_group} \leftarrow \mathtt{G}$

\vspace{0.2em}

\FOR{\tikzmarknode{a1}{} $\mathtt{n\_group} = 2$ \textbf{to} $\mathtt{G}$}

\vspace{0.2em}

\STATE $\mathtt{dp}[0,\mathtt{n\_group}] \leftarrow \mathtt{uni\_nchs}[0] * \mathtt{cum\_counts}[0]$

\vspace{0.2em}

\STATE $\mathtt{backtrace}[0,\mathtt{n\_group}] \leftarrow 0$

\vspace{0.2em}

\FOR{\tikzmarknode{a2}{} $\mathtt{i} = 1$ \textbf{to} $\mathtt{L} - 1$}

\vspace{0.2em}

\STATE $\mathtt{min\_overhead}, \mathtt{best\_idx} \leftarrow \mathtt{inf}, -1$

\vspace{0.2em}

\FOR{\tikzmarknode{a3}{} $\mathtt{j} = 0$ \textbf{to} $\mathtt{i} - 1$}

\vspace{0.2em}

\STATE $\mathtt{curr\_overhead} \leftarrow \mathtt{dp}[\mathtt{j}, \mathtt{n\_group} - 1] + \mathtt{uni\_nchs}[\mathtt{i}] * (\mathtt{cum\_counts}[\mathtt{i}] - \mathtt{cum\_counts}[\mathtt{j}])$

\vspace{0.2em}

\IF{\tikzmarknode{a4}{} $\mathtt{curr\_overhead} < \mathtt{min\_overhead}$}

\vspace{0.2em}

\STATE $\mathtt{min\_overhead}, \mathtt{best\_idx} \leftarrow \mathtt{curr\_overhead}, \mathtt{j}$

\vspace{0.2em}

\ENDIF

\ENDFOR

\STATE $\mathtt{dp}[\mathtt{i}, \mathtt{n\_group}], \mathtt{backtrace}[\mathtt{i}, \mathtt{n\_group}] \leftarrow \mathtt{min\_overhead}, \mathtt{best\_idx}$

\vspace{0.4em}

\ENDFOR

\IF{\tikzmarknode{a6}{} $\mathtt{dp}[-1,\mathtt{n\_group}] <= \mathtt{target\_overhead}$}

\vspace{0.2em}

\STATE $\mathtt{target\_n\_group} \leftarrow \mathtt{n\_group}$

\vspace{0.2em}

\ENDIF

\ENDFOR

\STATE \textcolor[RGB]{115,119,123}{\# Backtrace}

\vspace{0.2em}

\STATE $\mathtt{group\_sizes} \leftarrow (0)_{\mathtt{target\_n\_group}} \in \mathbb{Z}^{\mathtt{target\_n\_group}}$

\vspace{0.2em}

\STATE $\mathtt{i} \leftarrow \mathtt{L} - 1$

\vspace{0.2em}

\FOR{\tikzmarknode{a5}{} $n = \mathtt{target\_n\_group}$ \textbf{to} $1$}

\vspace{0.2em}

\STATE $\mathtt{group\_sizes}[n-1] \leftarrow \mathtt{i}$

\vspace{0.2em}

\STATE $\mathtt{i} \leftarrow \mathtt{backtrace}[\mathtt{i}, \mathtt{target\_n\_group}]$

\ENDFOR

\vspace{0.2em}

\STATE \textbf{return} $\mathtt{group\_sizes}$
    
\end{algorithmic}
}    
\end{algorithm}
\begin{tikzpicture}[overlay,remember picture]
    \draw[black,line width=0.6pt] ([xshift=-12pt,yshift=-3pt]a0.west) -- ([xshift=-12pt,yshift=-12pt]a0.west) -- ([xshift=-8pt,yshift=-12pt]a0.west);
    \draw[black,line width=0.6pt] ([xshift=-12pt,yshift=-3pt]a1.west) -- ([xshift=-12pt,yshift=-122pt]a1.west) -- ([xshift=-8pt,yshift=-122pt]a1.west);
    \draw[black,line width=0.6pt] ([xshift=-12pt,yshift=-3pt]a2.west) -- ([xshift=-12pt,yshift=-68pt]a2.west) -- ([xshift=-8pt,yshift=-68pt]a2.west);
    \draw[black,line width=0.6pt] ([xshift=-12pt,yshift=-3pt]a3.west) -- ([xshift=-12pt,yshift=-34pt]a3.west) -- ([xshift=-8pt,yshift=-34pt]a3.west);\draw[black,line width=0.6pt] ([xshift=-6pt,yshift=-3pt]a4.west) -- ([xshift=-6pt,yshift=-12pt]a4.west) -- ([xshift=-2pt,yshift=-12pt]a4.west);
    \draw[black,line width=0.6pt] ([xshift=-12pt,yshift=-3pt]a5.west) -- ([xshift=-12pt,yshift=-24pt]a5.west) -- ([xshift=-8pt,yshift=-24pt]a5.west);
    \draw[black,line width=0.6pt] ([xshift=-6pt,yshift=-3pt]a6.west) -- ([xshift=-6pt,yshift=-12pt]a6.west) -- ([xshift=-2pt,yshift=-12pt]a6.west);
\end{tikzpicture}
\vspace{-3.0em}
\end{figure}

We use a dynamic programming algorithm that is based on the following main idea. We first sort the numbers in $\mathtt{nchs}$ in ascending order. Denote $\mathtt{L}$ as the size of $\mathtt{nchs}$, we maintain a scratch table of size $\mathtt{L} \times \mathtt{G}$ whose $i$th row and $j$th column indicates the best possible overhead achieved by the first $i$ numbers in $\mathtt{nchs}$ when having in total at most $j$ partitions. The update formula of the DP table is
    \begin{align}
        \mathtt{dp}[i,j] \leftarrow \min_{k \in [i-1]} \mathtt{dp}[k,j-1] + \mathtt{nchs}[i] \cdot (i - k),
        \label{eq:dp-recur}
    \end{align}
\noindent where we try to find the best place ($k$) to put a new group/partition. By simultaneously maintaining a matrix for backtracking, we can retrieve the best partition found by the algorithm. 

The algorithm is shown in \cref{alg:layer-partitioning}. A practical trick to speed it up is to coalesce the identical values in $\mathtt{nchs}$ as done in line 3. Lines 7-9 initialize the buffers, and lines 11-23 are the main loop of the DP algorithm. Finally, the result partitioning is retrieved using lines 25-29.

\boldparagraph{Theoretical guarantee.} \cref{alg:layer-partitioning} is guaranteed to find an optimal grouping given a pre-specified number of groups, and is fairly efficient in practice. We formally state the problem in the following and provide the proof and analysis as follows.

As described in Appendix A.1, the grouping algorithm essentially takes as input a list of ``\# child node blocks'' for each parent node block in a layer, and the goal is to partition all parent node blocks into $K$ groups such that we minimize the following cost: the sum of the cost of each group, where the cost of a group is the maximum ``\# child node blocks'' in the group times the number of parent node blocks in the group. In the following, we first demonstrate that the proposed dynamic programming (DP) algorithm (Algorithm 2) can retain the optimal cost for every $K$. We then proceed to analyze the time and space complexity of the algorithm.

To simplify notations, we assume the input is a vector of integers $[n_1, \dots, n_N]$. We assume without loss of generality that the numbers are sorted because if not, we can apply any sorting algorithm. The main idea of the DP algorithm is to maintain a table termed dp of size N times K, where $\mathtt{dp}[i,j]$ indicates the optimal cost when partitioning the first i integers into j groups. For the base cases, we can set $\mathtt{dp}[i,1] = n_i (\forall i)$ and $\mathtt{dp}[1,j] = n_1 (\forall j)$. For the inductive case, we have \cref{eq:dp-recur}. It is straightforward to verify that when $\mathtt{dp}[k,j-1] (\forall k \in [1,i-1])$ are optimal, $\mathtt{dp}[i,j]$ is also optimal. Therefore, for any $K$, \cref{alg:layer-partitioning} computes the optimal grouping strategy for $K$ groups.

\boldparagraph{Efficiency.} We then focus on the runtime. Given $N$ and $K$, \cref{alg:layer-partitioning} requires $\bigO(KN^2)$ runtime and $\bigO(KN)$ memory, which is undesired for large $N$ (in practice, we set $K$ to be smaller than 10). However, as demonstrated in \cref{alg:layer-partitioning} (line 3), we only need to enumerate through the unique values in $[n_1, \dots, n_N]$, which could potentially lower the computation cost significantly. Even when we are dealing with highly non-structured PCs, we can always round the numbers up to a minimum integer that is divisible by a small integer such as 10. This allows us to achieve a decent approximated solution with much less computation time.

\subsection{Details of the Backpropagation Algorithm for Sum Layers}
\label{sec:backprop-details}

\begin{figure}[H]
\vspace{-0.8em}
\begin{algorithm}[H]
\caption{Backward pass of a sum layer group \wrt parameters}
\label{alg:bk-pass-params}
{\fontsize{9}{9} \selectfont
\begin{algorithmic}[1]

\STATE {\bfseries Inputs:} log-probs of product nodes $\boldsymbol{l}_{\mathrm{prod}}$, log-probs of sum nodes $\boldsymbol{l}_{\mathrm{sum}}$, flows of sum nodes $\mathtt{f}_{\mathrm{sum}}$, flattened parameter vector $\boldsymbol{\theta}_{\mathrm{flat}}$, $\mathtt{sum\_ids}$, $\mathtt{prod\_ids}, \mathtt{param\_ids}$

\vspace{0.2em}

\STATE {\bfseries Inputs:} \# sum nodes: $M$, \# product nodes: $N$, batch size: $B$

\vspace{0.2em}

\STATE {\bfseries Inputs:} block sizes $K_{\!M}$, $K_{\!N}$, $K_{\!B}$ for the sum node, product node, and batch dimensions, respectively

\vspace{0.2em}

\STATE {\bfseries Inputs:} number of sum node blocks $C_{\!M}$; number of product node blocks $C_{\!N}$; number of batch blocks $C_{\!B}$

\vspace{0.2em}

\STATE {\bfseries Outputs:} flows of params $\mathtt{f}_{\mathrm{params}}$

\vspace{0.2em}

\STATE {\bfseries Kernel launch:} schedule to launch $C_{\!M} \times C_{\!N}$ thread-blocks with $\mathtt{m} \!=\! 0, \dots, C_{\!M} \!-\! 1$ and $\mathtt{n} \!=\! 0, \dots, C_{\!N} \!-\! 1$

\vspace{0.2em}

\STATE $\mathtt{cum} \leftarrow (0)_{K_{\!M} \!\times\! K_{\!N}} \!\!\in \mathbb{R}^{K_{\!M} \times K_{\!N}}$ \hfill \textcolor[RGB]{115,119,123}{$\triangleright$ Scratch space on SRAM}

\vspace{0.2em}

\STATE $\mathtt{ms}, \mathtt{me} \leftarrow \mathtt{sum\_ids}[\mathtt{m}], \mathtt{sum\_ids}[\mathtt{m}] + K_{\!M}$

\vspace{0.2em}

\STATE $\mathtt{ns}, \mathtt{ne} \leftarrow \mathtt{prod\_ids}[\mathtt{m}, \mathtt{n}], \mathtt{prod\_ids}[\mathtt{m}, \mathtt{n}] + K_{\!N}$

\vspace{0.2em}

\FOR{\tikzmarknode{a1}{} $\! \mathtt{b} = 0$ \textbf{to} $C_{\!B} \!-\! 1$}

\vspace{0.2em}

\STATE $\mathtt{bs}, \mathtt{be} \leftarrow \mathtt{b} \cdot K_{\!B}, (\mathtt{b} + 1) \cdot K_{\!B}$ \hfill \textcolor[RGB]{115,119,123}{$\triangleright$ Start and end batch index}

\vspace{0.2em}

\STATE Load $\boldsymbol{f}_{\mathrm{s}} \!\leftarrow\! \mathtt{f}_{\mathrm{sum}}[\mathtt{ms} \!:\! \mathtt{me}, \mathtt{bs} \!:\! \mathtt{be}] \in \mathbb{R}^{K_{\!M} \times K_{\!B}}$ and $\boldsymbol{l}_{\mathrm{s}} \!\leftarrow\! \boldsymbol{l}_{\mathrm{sum}}[\mathtt{ms} \!:\! \mathtt{me}, \mathtt{bs} \!:\! \mathtt{be}] \in \mathbb{R}^{K_{\!M} \times K_{\!B}}$ to SRAM

\vspace{0.2em}

\STATE Load $\boldsymbol{l}_{\mathrm{p}} \!\leftarrow\! 
\boldsymbol{l}_{\mathrm{prod}}[\mathtt{ns} \!:\! \mathtt{ne},\mathtt{bs} \!:\! \mathtt{be}] \!\in\! \mathbb{R}^{K_{\!N} \times K_{\!B}}$ to SRAM

\vspace{0.2em}

\STATE $\mathtt{log\_nf} \leftarrow \log (\boldsymbol{f}_{\mathrm{s}}) - \boldsymbol{l}_{\mathrm{s}}$

\vspace{0.2em}

\STATE $\mathtt{log\_nf\_max} \leftarrow \mathtt{max}(\mathtt{log\_nf}, \mathtt{dim}\!=\!0) \in \mathbb{R}^{1 \times K_{\!B}}$ \hfill \textcolor[RGB]{115,119,123}{$\triangleright$ Compute on chip}

\vspace{0.2em}

\STATE $\mathtt{log\_nf\_sub} \leftarrow \mathtt{exp}(\mathtt{log\_nf} - \mathtt{log\_nf\_max}) \in \mathbb{R}^{K_{\!M} \times K_{\!B}}$

\vspace{0.2em}

\STATE $\mathtt{scaled\_emars} \leftarrow \mathtt{transpose}(\mathtt{exp} (\boldsymbol{\p}_{\mathrm{p}} + \mathtt{log\_nf\_max})) \in \mathbb{R}^{K_{\!B} \times K_{\!N}}$

\vspace{0.2em}

\STATE $\mathtt{partial\_flows} \leftarrow \mathtt{matmul}(\mathtt{log\_nf\_sub}, \mathtt{scaled\_emars}) \in \mathbb{R}^{K_{\!M} \times K_{\!N}}$ \hfill \textcolor[RGB]{115,119,123}{$\triangleright$ With Tensor Cores}

\vspace{0.2em}

\STATE $\mathtt{cum} \leftarrow \mathtt{cum} + \mathtt{partial\_flows}$

\ENDFOR

\vspace{0.2em}

\STATE $\mathtt{ps}, \mathtt{pe} \leftarrow \mathtt{param\_ids}[\mathtt{m},\mathtt{n}], \mathtt{param\_ids}[\mathtt{m},\mathtt{n}] + K_{\!M} \cdot K_{\!N}$

\vspace{0.2em}

\STATE $\mathtt{f}_{\mathrm{params}}[\mathtt{ps} \!:\! \mathtt{pe}] \leftarrow \mathtt{f}_{\mathrm{params}}[\mathtt{ps} \!:\! \mathtt{pe}] +  \boldsymbol{\theta}_{\mathrm{flat}}[\mathtt{ps} \!:\! \mathtt{pe}] * \mathtt{cum}.\mathtt{view}(K_{\!M} * K_{\!N})$

\end{algorithmic}
}    
\end{algorithm}
\begin{tikzpicture}[overlay,remember picture]
    \draw[black,line width=0.6pt] ([xshift=-12pt,yshift=-3pt]a1.west) -- ([xshift=-12pt,yshift=-112pt]a1.west) -- ([xshift=-8pt,yshift=-112pt]a1.west);
\end{tikzpicture}
\vspace{-4.0em}
\end{figure}

We compute the backward pass with respect to the inputs and the parameters of the sum layer in two different kernels as we need two different layer partitioning strategies to improve efficiency. In the following, we first introduce the backpropagation algorithm for the parameters since it reuses the index tensors compiled for the forward pass (\ie $\mathtt{sum\_ids}$, $\mathtt{prod\_ids}$, and $\mathtt{param\_ids}$).

The algorithm is shown in \cref{alg:bk-pass-params}. In addition to the log-probabilities of the product nodes (\ie $\boldsymbol{l}_{\mathrm{prod}}$), the log-probabilities of the sum nodes (\ie $\boldsymbol{l}_{\mathrm{sum}}$), and the flattened parameters (\ie $\boldsymbol{\theta}_{\mathrm{flat}}$), the algorithm takes as input the flows $\mathtt{f}_{\mathrm{sum}}$ computed for the sum nodes. Following \cref{def:pc-flows}, we can compute the flow \wrt the sum parameters as 
    \begin{align*}
        \mathrm{F}_{n,c} (\x) := \theta_{n,c} \cdot \p_{c} (\x) / \p_{n} (\x) \cdot \mathrm{F}_{n} (\x).
    \end{align*}
Similar to \cref{alg:fwd-pass}, we partition the sum nodes, product nodes, and samples into blocks of size $K_{\!M}$, $K_{\!N}$, and $K_{\!B}$, respectively. We schedule to launch $C_{\!M} \!\times\! C_{\!N}$ thread-blocks, each responsible for computing the parameter flows for a block of $K_{\!M} \!\times\! K_{\!N}$ parameter flows. The main loop (line 10) iterates through blocks of $K_{\!B}$ samples. In every iteration, we first load the log-probabilities (\ie $\boldsymbol{l}_{\mathrm{s}}$ and $\boldsymbol{l}_{\mathrm{p}}$) and the sum node flows (\ie $\boldsymbol{f}_{\mathrm{s}}$) to compute the partial flow $\p_{c} (\x) / \p_{n} (\x) \cdot \mathrm{F}_{n} (\x)$ for the block of samples (note that this equals $\mathrm{F}_{n,c} (\x) / \theta_{n,c}$. The partial flows are accumulated in the matrix $\mathtt{cum}$ initialized in line 7. After processing all blocks of samples, we add back the parameter flows by accumulating $\mathtt{cum} * [\text{the~corresponding~parameters}]$ in line 21.

As elaborated in \cref{sec:eff-backward}, if we use the same set of index tensors used in the forward pass, we have the problem of different thread-blocks needing to write (partial) flows to the same input product node blocks. Therefore, we do a separate compilation step for the backward pass. Consider a sum layer with sum node blocks of size $K_{\!M}$ and child product node blocks of size $K_{\!N}$. We first partition the $C_{\!N}$ children into groups such that every child node block in a group has a similar number of parents. This is done by the dynamic programming algorithm described in \cref{sec:layer-partitioning}.

Similar to the compilation procedure of the forward pass, for a group with $C_{\!N}$ child node blocks (assume every block has $C_{\!M}$ blocks of parents), we generate three index tensors: $\mathtt{ch\_ids} \!\in\! \mathbb{Z}^{C_{\!N}}$ and $\mathtt{par\_ids}, \mathtt{par\_param\_ids} \!\in\! \mathbb{Z}^{C_{\!N} \times C_{\!M}}$. $\mathtt{ch\_ids}$ contains the initial index of all $C_{\!N}$ child node blocks belonging to the group. For the $i$th node block in the group (\ie the product node block with the initial index $\mathtt{ch\_ids}[i]$), $\mathtt{par\_ids}[i,:]$ encode the start indices of its parent sum node blocks, and $\mathtt{par\_param\_ids}[i,:]$ represent the corresponding initial parameter indices.

The main algorithmic procedure is very similar to $\cref{alg:fwd-pass}$. Specifically, the kernel schedules to launch $C_{\!N} \!\times\! C_{\!B}$ thread-blocks each computing a block of $K_{\!N} \!\times\! K_{\!B}$ product node flows. In the main loop (line 9), we iterate through all $C_{\!M}$ parent node blocks. In lines 13-16, we are essentially computing $\theta_{n,c} / \p_{n} (\x) \cdot \mathrm{F}_{n} (\x)$ (notations inherited from \cref{def:pc-flows}) for the block of $K_{\!N} \!\times\! K_{\!B}$ values using the logsumexp trick. Finally, we store the results back to $\mathtt{f}_\mathrm{prod}$.

\begin{figure}[h]
\vspace{-0.8em}
\begin{algorithm}[H]
\caption{Backward pass of a sum layer group \wrt inputs}
\label{alg:bk-pass-inputs}
{\fontsize{9}{9} \selectfont
\begin{algorithmic}[1]

\STATE {\bfseries Inputs:} log-probs of product nodes $\boldsymbol{l}_{\mathrm{prod}}$, log-probs of sum nodes $\boldsymbol{l}_{\mathrm{sum}}$, flows of sum nodes $\mathtt{f}_{\mathrm{sum}}$, flattened parameter vector $\boldsymbol{\theta}_{\mathrm{flat}}$, $\mathtt{ch\_ids}$, $\mathtt{par\_ids}, \mathtt{par\_param\_ids}$

\vspace{0.2em}

\STATE {\bfseries Inputs:} \# sum nodes: $M$, \# product nodes: $N$, batch size: $B$

\vspace{0.2em}

\STATE {\bfseries Inputs:} block sizes $K_{\!M}$, $K_{\!N}$, $K_{\!B}$ for the sum node, product node, and batch dimensions, respectively

\vspace{0.2em}

\STATE {\bfseries Inputs:} number of sum node blocks $C_{\!M}$; number of product node blocks $C_{\!N}$; number of batch blocks $C_{\!B}$

\vspace{0.2em}

\STATE {\bfseries Outputs:} flows of inputs $\mathtt{f}_{\mathrm{prod}}$

\vspace{0.2em}

\STATE {\bfseries Kernel launch:} schedule to launch $C_{\!N} \times C_{\!B}$ thread-blocks with $\mathtt{n} \!=\! 0, \dots, C_{\!N} \!-\! 1$ and $\mathtt{b} \!=\! 0, \dots, C_{\!B} \!-\! 1$

\vspace{0.2em}

\STATE $\mathtt{cum} \leftarrow (-\infty)_{K_{\!N} \!\times\! K_{\!B}} \!\!\in \mathbb{R}^{K_{\!N} \times K_{\!B}}$ \hfill \textcolor[RGB]{115,119,123}{$\triangleright$ Scratch space on SRAM}

\vspace{0.2em}

\STATE $\mathtt{bs}, \mathtt{be} \leftarrow \mathtt{b} \cdot K_{\!B}, (\mathtt{b} + 1) \cdot K_{\!B}$

\vspace{0.2em}

\FOR{\tikzmarknode{a1}{} $\! \mathtt{m} = 0$ \textbf{to} $C_{\!M} \!-\! 1$}

\vspace{0.2em}

\STATE $\mathtt{ps}, \mathtt{pe} \leftarrow \mathtt{par\_param\_ids}[\mathtt{n},\mathtt{m}]$

\vspace{0.2em}

\STATE Load $\boldsymbol{f}_{\mathrm{s}} \!\leftarrow\! \mathtt{f}_{\mathrm{sum}}[\mathtt{ms} \!:\! \mathtt{me}, \mathtt{bs} \!:\! \mathtt{be}] \in \mathbb{R}^{K_{\!M} \times K_{\!B}}$ and $\boldsymbol{l}_{\mathrm{s}} \!\leftarrow\! \boldsymbol{l}_{\mathrm{sum}}[\mathtt{ms} \!:\! \mathtt{me}, \mathtt{bs} \!:\! \mathtt{be}] \in \mathbb{R}^{K_{\!M} \times K_{\!B}}$ to SRAM

\vspace{0.2em}

\STATE Load $\boldsymbol{\theta} \!\leftarrow\! 
\mathtt{transpose}(\boldsymbol{\theta}_{\mathrm{flat}}[\mathtt{ps} \!:\! \mathtt{pe}].\mathtt{view}(K_{\!M}, K_{\!N})) \!\in\! \mathbb{R}^{K_{\!N} \times K_{\!M}}$ to SRAM

\vspace{0.2em}

\STATE $\mathtt{log\_nf} \leftarrow \log (\boldsymbol{f}_{\mathrm{s}}) - \boldsymbol{l}_{\mathrm{s}}$

\vspace{0.2em}

\STATE $\mathtt{log\_nf\_max} \leftarrow \mathtt{max}(\mathtt{log\_nf}, \mathtt{dim}\!=\!0) \in \mathbb{R}^{1 \times K_{\!B}}$ \hfill \textcolor[RGB]{115,119,123}{$\triangleright$ Compute on chip}

\vspace{0.2em}

\STATE $\mathtt{log\_nf\_sub} \leftarrow \mathtt{exp}(\mathtt{log\_nf} - \mathtt{log\_nf\_max}) \in \mathbb{R}^{K_{\!M} \times K_{\!B}}$

\vspace{0.2em}

\STATE $\mathtt{partial\_flows} \leftarrow \mathtt{matmul}(\boldsymbol{\theta}, \mathtt{log\_nf\_sub}) \in \mathbb{R}^{K_{\!M} \times K_{\!N}}$ \hfill \textcolor[RGB]{115,119,123}{$\triangleright$ With Tensor Cores}

\vspace{0.2em}

\STATE \hspace{-0.5em}
    {\setlength{\abovedisplayskip}{-0.2em}
    \setlength{\belowdisplayskip}{0.2em}
    \hspace*{-11em}\vbox{\begin{align*}
    \fontsize{9}{9}\selectfont
    \mathtt{cum} \leftarrow \mathtt{where}( & \mathtt{log\_nf\_max} > \mathtt{cum}, \\[-0.1em]
    & \mathtt{log}(\mathtt{partial\_flows} + \mathtt{exp}(\mathtt{cum} - \mathtt{log\_nf\_max}) + \mathtt{log\_nf\_max}, \\[-0.1em]
    & \mathtt{log}(\mathtt{exp}(\mathtt{log\_nf\_max} - \mathtt{cum}) \cdot \mathtt{partial\_flows} + 1) + \mathtt{cum})
\end{align*}}}

\ENDFOR

\vspace{0.2em}

\STATE $\mathtt{ns}, \mathtt{ne} \leftarrow \mathtt{ch\_ids}[\mathtt{n}], \mathtt{ch\_ids}[\mathtt{n}] + K_{\!N}$

\vspace{0.2em}

\STATE $\mathtt{f}_{\mathrm{prod}}[\mathtt{ns} \!:\! \mathtt{ne}, \mathtt{bs} \!:\! \mathtt{be}] \leftarrow \exp(\mathtt{cum} + \boldsymbol{l}_{\mathrm{prod}}[\mathtt{ns} \!:\! \mathtt{ne}, \mathtt{bs} \!:\! \mathtt{be}])$

\end{algorithmic}
}    
\end{algorithm}
\begin{tikzpicture}[overlay,remember picture]
    \draw[black,line width=0.6pt] ([xshift=-12pt,yshift=-3pt]a1.west) -- ([xshift=-12pt,yshift=-126pt]a1.west) -- ([xshift=-8pt,yshift=-126pt]a1.west);
\end{tikzpicture}
\vspace{-2.0em}
\end{figure}

\subsection{PCs with Tied Parameters}
\label{sec:param-tying}

Formally, PCs with tied parameters are PCs containing same sub-structures in different parts of its DAG. Although the nodes in these sub-structures could have different semantics, they can have shared/tied parameters. For example, in homogeneous HMMs, although the transition probabilities between different pairs of consecutive latent variables are represented by different sets of nodes and edges in the PC, they all have the \emph{same} set of probability parameters.

PyJuice can be readily adapted to PCs with tied parameters. For the forward pass, we just need the compiler to assign the same parameter indices in $\mathtt{param\_ids}$. Similarly, we only need to slightly change the compilation procedure of $\mathtt{par\_param\_ids}$. One notable difference is that in the backward pass \wrt the parameters, multiple thread-blocks would need to write partial flows to the same memory addresses, which leads to inter-thread-block barriers. We implemented a memory-IO tradeoff by letting the compiler create new sets of memory addresses to store the parameter flows when the number of thread-blocks writing to the same address is greater than a predefined threshold (by default set to $4$).

\section{Additional Technical Details}

\subsection{Block-Sparsity of Common PC Structures}
\label{sec:pc-block-sparsity}

Most commonly-adopted PC structures such as PD \citep{poon2011sum}, RAT-SPN \citep{peharz2020random}, and HCLT \citep{liu2021tractable} have block-sparse sum layers because one of the key building blocks of the structure is a set of sum nodes fully connected to their inputs. Therefore, every sum layer must contain multiple fully-connected blocks of sum and product nodes, and hence they are block sparse.

\subsection{Relation Between PC Flows and Gradients}
\label{sec:flow-grad-rel}

We first show the equality for the node flows: 
    \begin{align}
        \mathrm{F}_{n} (\x) = \frac{\partial \log \p_{n_r} (\x)}{\partial \log \p_{n} (\x)}.
        \label{eq:node-flow-eq}
    \end{align}
We do the proof by induction. As a base case, we have by definition that $\mathrm{F}_{n_r} (\x) = \partial \log \p_{n_r} (\x) / \partial \log \p_{n_r} (\x) = 1$.

Next, suppose $n$ is a sum or an input node, and for all its parents $m$, we have \cref{eq:node-flow-eq} is satisfied by induction. Since all parents of $n$ are product nodes, we have
    \begin{align*}
        \mathrm{F}_{n} (\x) = \sum_{m \in \pa(n)} \mathrm{F}_{m} (\x) = \sum_{m \in \pa(n)} \frac{\partial \log \p_{n_r} (\x)}{\partial \log \p_{m} (\x)} = \sum_{m \in \pa(n)} \frac{\partial \log \p_{n_r} (\x)}{\partial \log \p_{n \rightarrow m} (\x)} = \frac{\partial \log \p_{n_r} (\x)}{\partial \log \p_{n} (\x)},
    \end{align*}
\noindent where $\p_{n \rightarrow m} (\x)$ denotes the probability carried by the edge from $n$ to $m$.

Finally, suppose $n$ is a product node and thus all its parents are sum nodes. We have
    \begin{align}
        \mathrm{F}_{n} (\x) & = \sum_{m \in \pa(n)} \frac{\theta_{m,n} \cdot \p_{n} (\x)}{\p_{m} (\x)} \cdot \mathrm{F}_{m} (\x) = \sum_{m \in \pa(n)} \frac{\theta_{m,n} \cdot \p_{n} (\x)}{\p_{m} (\x)} \cdot \frac{\partial \log \p_{n_r} (\x)}{\partial \log \p_{m} (\x)}, \\
        & = \sum_{m \in \pa(n)} \theta_{m,n} \cdot \p_{n} (\x) \cdot \frac{\partial \log \p_{n_r} (\x)}{\partial \p_{m} (\x)}.
        \label{eq:proof-1}
    \end{align}
Denote $\p_{n \rightarrow m} (\x) = \theta_{m,n} \cdot \p_{n} (\x)$ as the probability carried on the edge $(m,n)$. Since $\p_{m} (\x) = \sum_{n' \in \ch(m)} \p_{n' \rightarrow m} (\x)$, we have 
    \begin{align*}
        \forall n \in \ch(m), \frac{\partial \log \p_{n_r} (\x)}{\partial \p_{m} (\x)} = \frac{\partial \log \p_{n_r} (\x)}{\partial \p_{n \rightarrow m} (\x)}.
    \end{align*}
Plug in the above equation on $\mathrm{F}_{n} (\x)$, this results in
    \begin{align}
        \mathrm{F}_{n} (\x) = \sum_{m \in \pa(n)} \p_{n \rightarrow m} (\x) \cdot \frac{\partial \log \p_{n_r} (\x)}{\partial \p_{n \rightarrow m}(\x)} = \sum_{m \in \pa(n)} \frac{\partial \log \p_{n_r} (\x)}{\partial \log \p_{n \rightarrow m}(\x)} = \frac{\partial \log \p_{n_r} (\x)}{\partial \log \p_{n} (\x)}.
        \label{eq:proof-2}
    \end{align}

We move on to demonstrate the following relation:
    \begin{align*}
        \mathrm{F}_{n,c} (\x) = \theta_{n,c} \cdot \frac{\partial \log \p_{n_r} (\x)}{\partial \theta_{n,c}} = \frac{\partial \log \p_{n_r} (\x)}{\partial \log \theta_{n,c}},
    \end{align*}
\noindent where $n$ is a sum node and $c$ is one of its children. We reuse the results derived in \cref{eq:proof-1,eq:proof-2}, where we replace $n$ with $c$ and $m$ with $n$:
    \begin{align*}
        \mathrm{F}_{n,c} (\x) = \frac{\theta_{n,c} \cdot \p_{c} (\x)}{\p_{n} (\x)} \cdot \mathrm{F}_{n} (\x) = \theta_{n,c} \cdot \p_{c} (\x) \cdot \frac{\partial \log \p_{n_r} (\x)}{\partial \p_{n} (\x)} = \frac{\partial \log \p_{n_r} (\x)}{\partial \log \p_{c \rightarrow n} (\x)} = \frac{\partial \log \p_{n_r} (\x)}{\partial \log \theta_{n,c}}.
    \end{align*}

% \section{Additional Related Work}

% \guy{Can we save the pluging for later and here try to make the most compelling case quickly?}

\section{Experimental Details}
\label{sec:exp-details}

\subsection{The Adopted Block-Sparse PC Layer}
\label{sec:exp-blk-sp-layer}

The PC layer contains $200$ independent fully-connected sets of nodes. Every connected subset consists of $1024$ sum nodes and $1024$ product nodes. When compiling the layer, we divide the layer into blocks of size $32$. When dropping $32 \!\times\! 32$ edge blocks from the layer, we ensure that every sum node has at least one child.

\subsection{Details of Training the HMM Language Model}
\label{sec:exp-gelato-hmm}

Following \citet{zhang2023tractable}, we first fine-tune a GPT-2 model with the CommonGen dataset. We then sample 8M sequences of length $32$ from the fine-tuned GPT-2. After initializing the HMM parameters with latent variable distillation, we fine-tune the HMM with the sampled data. Specifically, following \citet{zhang2023tractable}, we divide the 8M samples into $40$ equally-sized subsets, and run full-batch EM on the 40 subsets repeatedly. Another set of 800K samples is drawn from the fine-tuned GPT as the validation set.

\subsection{Details of Training the Sparse Image Model}
\label{sec:exp-lvd-pg}

Following \citet{liu2023understanding}, we fine-tune the model with an equivalent batch size of $6400$ and a step size of $0.01$ in the mini-batch EM algorithm. Specifically, suppose $\boldsymbol{\theta}$ are the current parameters, $\boldsymbol{\theta}^{\mathrm{new}}$ are the new set of parameters computed by the EM update. Given step size $\alpha$, the update formula is $\boldsymbol{\theta} \leftarrow (1 - \alpha) \boldsymbol{\theta} + \alpha \boldsymbol{\theta}^{\mathrm{new}}$.

\subsection{Additional Benchmark Results}
\label{sec:exp-benchmark-details}

\boldparagraph{Hyperparameters of the adopted HCLTs.}
We adopt two HCLTs \citep{liu2021tractable} with hidden sizes $256$ and $512$, respectively. The backbone CLT structure is constructed using 20,000 randomly selected training samples.

\boldparagraph{Hyperparameters of the adopted PDs.}
Starting from the set of all random variables, the PD structure recursively splits the subset with product nodes. Specifically, consider an image represented as a $H \!\times\! W \!\times\! C$ ($H$ is the hight; $W$ is the width; $C$ is the number of channels), the PD structure recursively splits over both the height and the width coordinates, where every coordinate has a set of pre-defined split points. For both the height and the width coordinates, we add split points with interval $2$. PD-mid has a hidden dimension of $128$ and PD-large has $256$.

\boldparagraph{Benchmark results on WikiText-103.}
\cref{tab:benchmarking-wikitext} illustrates results on WikiText-103. We train the model on sequences with $64$ tokens. We adopt two (homogeneous) HMM models, HMM-mid and HMM-large with hidden sizes $2048$ and $4096$, respectively.

\begin{table}[H]
    \centering
    \vspace{-1.2em}
    \caption{Density estimation performance of PCs on the WikiText-103 dataset. Reported numbers are test set perplexity.}
    \label{tab:benchmarking-wikitext}
    \vspace{0.2em}
    \renewcommand{\arraystretch}{0.88}
    \centering
    \scalebox{0.84}{
    \begin{tabular}{lcc}
        \toprule
        Dataset & HMM-mid & HMM-large \\
        \midrule
        WikiText-103 & 146.59 & 167.65 \\
        \bottomrule
    \end{tabular}}
\vspace{-1.4em}
\end{table}

\section{Additional Experiments}

\subsection{Speed of the Compilation Process}
\label{sec:exp-compilation-speed}

In \cref{tab:compilation-speed}, we show the compilation speed of PCs with different structures and different sizes. Experiments are conducted on a server with an AMD EPYC 7763 64-Core Processor and 8 RTX 4090 GPUs (we only use one GPU). The results demonstrate the efficiency of the compilation process, where even the PD model with close to 1B parameters can be compiled in around $30$ seconds.

\begin{table}[H]
    \centering
    \vspace{-1.2em}
    \caption{Average ($\pm$ standard deviation of 3 runs) runtime (in seconds) of the compilation process of four PCs.}
    \label{tab:compilation-speed}
    \vspace{0.2em}
    \renewcommand{\arraystretch}{0.88}
    \centering
    \scalebox{0.84}{
    \begin{tabular}{lMMMM}
        \toprule
        Structure & \multicolumn{2}{c}{HMM} & \multicolumn{2}{c}{PD} & \multicolumn{2}{c}{HCLT} & \multicolumn{2}{c}{RAT-SPN} \\
        \midrule
        \# nodes & 130&K & 1.38&M & 710&K & 465&K \\
        \# edges & 130&M & 829&M & 159&M & 33.4&M \\
        \midrule
        Compilation time (s) & 1.50&{\tiny $\pm$0.02} & 30.57&{\tiny $\pm$0.86} & 8.70&{\tiny $\pm$0.32} & 4.72&{\tiny $\pm$0.16} \\
        \bottomrule
    \end{tabular}}
\vspace{-1.4em}
\end{table}

\subsection{Runtime on Different GPUs}

In addition to the RTX 4090 GPU adopted in the experiments in \cref{tab:speed-results}, we compare the runtime of PyJuice with the baselines on an NVIDIA A40 GPU. As shown in the following table, PyJuice is still significantly faster than all baselines for PCs of different sizes.

\begin{table}[H]
    \centering
    \vspace{-1.2em}
    \caption{\textbf{Average ($\pm$ standard deviation of $5$ runs) runtime (in seconds) per training epoch} of 60K samples for PyJuice and the baselines on five RAT-SPNs \citep{peharz2020random} with different sizes. All other settings are the same as described in \cref{sec:exp-faster-pcs}.}
    \label{tab:speed-results-A40}
    \vspace{0.2em}
    \renewcommand{\arraystretch}{0.88}
    \centering
    \scalebox{0.84}{
    \begin{tabular}{lMMMMM}
        \toprule
        \# nodes & \multicolumn{2}{c}{58K} & \multicolumn{2}{c}{116K} & \multicolumn{2}{c}{232K} & \multicolumn{2}{c}{465K} & \multicolumn{2}{c}{930K} \\
        \# edges & \multicolumn{2}{c}{616K} & \multicolumn{2}{c}{2.2M} & \multicolumn{2}{c}{8.6M} & \multicolumn{2}{c}{33.4M} & \multicolumn{2}{c}{132M} \\
        \midrule
        EiNet & 60.29&{\tiny $\pm$0.30} & 136.85&{\tiny $\pm$0.13} & 282.58&{\tiny $\pm$0.27} & 690.73&{\tiny $\pm$0.08} & 1936.28&{\tiny $\pm$0.26} \\
        Juice.jl & 4.41&{\tiny $\pm$0.21} & 11.57&{\tiny $\pm$0.07} & 32.74&{\tiny $\pm$1.86} & 121.25&{\tiny $\pm$0.43} & 331.98&{\tiny $\pm$2.87} \\
        PyJuice & \textbf{1.53}&\tiny{$\pm$0.07} & \textbf{3.11}&\tiny{$\pm$0.07} & \textbf{6.47}&\tiny{$\pm$0.08} & \textbf{13.62}&\tiny{$\pm$0.37} & \textbf{30.69}&\tiny{$\pm$0.19} \\
        \bottomrule
    \end{tabular}}
\vspace{-1.4em}
\end{table}

\subsection{Runtime on Different Batch Sizes}

As a supplement to \cref{tab:speed-results}, we report the runtime for a RAT-SPN \citep{peharz2020random} with $465K$ nodes and $33.4M$ edges using batch sizes $\{8, 16, 32, 64, 128, 256, 512\}$. To minimize distractions, we only record the time to compute the forward and backward process, but not the time used for EM updates. Results are shown in the table below.

\begin{table}[H]
    \centering
    \vspace{-1.2em}
    \caption{\textbf{Average ($\pm$ standard deviation of $5$ runs) runtime (in seconds) per training epoch (excluding EM updates)} of 60K samples for PyJuice and the baselines on a RAT-SPNs \citep{peharz2020random} with $465K$ nodes and $33.4M$ edges. All other settings are the same as described in \cref{sec:exp-faster-pcs}. OOM denotes out-of-memory.}
    \label{tab:speed-results-batch}
    \vspace{0.2em}
    \renewcommand{\arraystretch}{0.88}
    \centering
    \scalebox{0.84}{
    \begin{tabular}{lMMMMMMM}
        \toprule
        Batch size & \multicolumn{2}{c}{8} & \multicolumn{2}{c}{16} & \multicolumn{2}{c}{32} & \multicolumn{2}{c}{64} & \multicolumn{2}{c}{128} & \multicolumn{2}{c}{256} & \multicolumn{2}{c}{512} \\
        \midrule
        EiNet & 332.87&{\tiny $\pm$0.21} & \multicolumn{2}{c}{OOM} & \multicolumn{2}{c}{OOM} & \multicolumn{2}{c}{OOM} & \multicolumn{2}{c}{OOM} & \multicolumn{2}{c}{OOM} & \multicolumn{2}{c}{OOM} \\
        Juice.jl & 1045.04&{\tiny $\pm$0.06} & 853.15&{\tiny $\pm$0.03} & 775.87&{\tiny $\pm$0.02} & 642.54&{\tiny $\pm$0.04} & 324.23&{\tiny $\pm$0.02} & 163.68&{\tiny $\pm$0.02} & 80.57&{\tiny $\pm$0.01} \\
        PyJuice & \textbf{43.09}&\tiny{$\pm$0.04} & \textbf{18.63}&\tiny{$\pm$0.02} & \textbf{7.38}&\tiny{$\pm$0.01} & \textbf{4.58}&\tiny{$\pm$0.01} & \textbf{3.50}&\tiny{$\pm$0.01} & \textbf{3.04}&\tiny{$\pm$0.01} & \textbf{2.76}&\tiny{$\pm$0.03} \\
        \bottomrule
    \end{tabular}}
\vspace{-1.4em}
\end{table}

%%%%%%%%%%%%%%%%%%%%%%%%%%%%%%%%%%%%%%%%%%%%%%%%%%%%%%%%%%%%%%%%%%%%%%%%%%%%%%%
%%%%%%%%%%%%%%%%%%%%%%%%%%%%%%%%%%%%%%%%%%%%%%%%%%%%%%%%%%%%%%%%%%%%%%%%%%%%%%%

\end{document}